%% file: main.tex
\documentclass{article}
\input{def.tex}

\usepackage{microtype}
\usepackage{graphicx}
\usepackage{algorithm}
\usepackage{algorithmic}

\allowdisplaybreaks

\title{Doubly Robust Off-Policy Value and Gradient Estimation for Deterministic Policies}
\author{Nathan Kallus\\
       Cornell University\\
       New York, NY 10044, USA
       \and
        Masatoshi Uehara\thanks{\url{uehara_m@g.harvard.edu}}\\
       Harvard University\\
       Cambridge, MA 02138, USA}

\date{}

\begin{document}

\maketitle

\begin{abstract}
Offline reinforcement learning, wherein one uses off-policy data logged by a fixed behavior policy to evaluate and learn new policies, is crucial in applications where experimentation is limited such as medicine. We study the estimation of policy value and gradient of a deterministic policy from off-policy data when actions are continuous. Targeting deterministic policies, for which action is a deterministic function of state, is crucial since optimal policies are always deterministic (up to ties). In this setting, standard importance sampling and doubly robust estimators for policy value and gradient fail because the density ratio does not exist. To circumvent this issue, we propose several new doubly robust estimators based on different kernelization approaches. We analyze the asymptotic mean-squared error of each of these under mild rate conditions for nuisance estimators. Specifically, we demonstrate how to obtain a rate that is independent of the horizon length.
\end{abstract}

\section{Introduction}

Offline reinforcement learning (RL), wherein one uses off-policy data logged by a fixed behavior policy to evaluate and learn new policies, is crucial in applications where experimentation is limited \citep{Chow2018,pmlr-v97-bibaut19a,NIPS2018_7530,Kallus2019IntrinsicallyES}. A key application is RL for healthcare \citep{gottesman2019guidelines,MurphyS.A.2003Odtr}. Since it is not possible to collect new data, it is crucial to efficiently use the available data. Recent work on off-policy evaluation \citep[OPE;][]{KallusUehara2019,KallusNathan2019EBtC} have shown how efficiently taking advantage of problem structure, such as Markovianness and ergodicity, can improve OPE and tackle the well-known issue of OPE known as the curse of horizon \citep{Liu2018}. 
\citet{EEOPG2020} applied these advances to off-policy \emph{learning} using a policy gradient approach, i.e., proposing efficient off-policy estimators for the policy gradient and incorporating them into gradient ascent methods.

All the aforementioned methods, however, cannot be directly applied to the evaluation and learning of deterministic policies when actions are continuous since the density ratio (Radon-Nikodym derivative) does not exist: the behavior policy usually has zero mass on single actions (more generally, it can have at most countably-many atoms while the evaluation policy may take any of a continuum of actions). This question is important since the maximum-value policy is generally deterministic (up to ties between actions) so if one seeks optimal policies one should focus on deterministic ones. In the bandit setting (horizon of one action), several recent works tackle this problem \citep{BibautAurelien2017Dsfo,pmlr-v84-kallus18a,Kyle2019},
but applying these methods in a straightforward manner to RL may lead to a bad convergence rate that deteriorates with horizon.

In this paper, we propose several doubly robust off-policy value and gradient estimators for deterministic policies in an RL setting. We analyze the asymptotic mean-squared error (MSE) of each estimator under extremely lax conditions that accommodate flexible learning of the nuisances that appear in the estimator (such as $q$-functions). 
Specifically, we propose estimators of policy value and gradient with MSE convergence rate that does not deteriorate with horizon and the leading term's coefficient has only a polynomial dependence on horizon.
These results are summarized in \cref{tab:all_compariosn}.

\begin{table}[t!]%
\centering 
\caption{Comparison of off-policy value and gradient estimators for deterministic policies. Proposed estimators are typeset in bold. ``MSE'' is the \emph{convergence rate} when nuisances are estimated at rate under ``Rate,'' \emph{irrespective} of choice of nuisance estimators. ``--'' means the convergence rate depends on the choice of estimators. ``Nuis'' are nuisances for OPE. ``Nuis+'' are \emph{additional} nuisances for policy gradient. ``D/I'' is whether differentiation (D) or integration (I) of $\hat q_t(s_t,a_t)$ wrt $a_t$ is required.}\label{tab:all_compariosn}
\begin{minipage}[t]{0.45\textwidth}\centering
\captionof{subtable}{OPE}
{\small
\begin{tabular}{llll}\toprule
&   MSE  &  Nuis & Rate  \\ \midrule
DM   & -- &  $q^{\epol}$\vphantom{$d^{q^{\epol}}$}   &   --\\
\textbf{CDRD} & $n^{-\frac{4}{H+4}}$  & $\lambda^{\Kcal},q^{\Kcal}$\vphantom{$d^{\lambda^{\Kcal}},d^{q^{\Kcal}}$}    &  $n^{-\frac{1}{H+4}}$ \\
\textbf{CDRK} &$n^{-\frac{4}{H+4}}$  & $\lambda^{\epol},q^{\epol}$\vphantom{$d^{\lambda^{\epol}},d^{q^{\epol}}$}   &  $n^{-\frac{1}{H+4}}$  \\
\textbf{MDRD} & $n^{-\frac{4}{5}}$ & $w^{\Kcal},q^{\Kcal}$\vphantom{$d^{w^{\Kcal}},d^{q^{\Kcal}}$}  &  $n^{-\frac{1}{5}}$ \\
\textbf{MDRK} & $n^{-\frac{4}{5}}$  & $w^{\epol},q^{\epol}$\vphantom{$d^{w^{\epol}},d^{q^{\epol}}$}  &  $n^{-\frac{1}{5}}$\\ \bottomrule
\end{tabular}
\label{tab:ope}
}
\end{minipage}%
\begin{minipage}[t]{0.55\textwidth}\centering
\captionof{subtable}{Off-policy gradient}
{\small
\begin{tabular}{lllll}\toprule
&  MSE  & Nuis+ &   Rate & D/I   \\ \midrule
DPG   &  --  &  $d^{q^{\epol}}$    &   --   & D    \\
\textbf{CPGD} & $n^{-\frac{4}{H+6}}$ & $d^{\lambda^{\Kcal}},d^{q^{\Kcal}}$    & $n^{-\frac{1}{H+6}}$& D  \\
\textbf{CPGK} &$n^{-\frac{4}{H+6}}$ & $d^{\lambda^{\epol}},d^{q^{\epol}}$   &  $n^{-\frac{1}{H+6}}$ &  I   \\
\textbf{MPGD} & $n^{-\frac{4}{7}}$ & $d^{w^{\Kcal}},d^{q^{\Kcal}}$  & $n^{-\frac{1}{7}}$ &  D \\
\textbf{MPGK} & $n^{-\frac{4}{7}}$  & $d^{w^{\epol}},d^{q^{\epol}}$  &  $n^{-\frac{1}{7}}$ &  I\\ \bottomrule
\end{tabular}
\label{tab:opg}
}\end{minipage}
\end{table}

\section{Preliminaries}

\paragraph{Problem set up} \label{sec:notationanddefs}

Consider an $H$-long time-varying Markov decision process (MDP), with states $s_t\in\mathcal S_t$, actions $a_t\in\Acal_t$, rewards $r_t\in[0,R_{\max}]$, initial state distribution $p_1(s_1)$, transition distributions $p_{t+1}(s_{t+1}\mid s_t,a_t)$, and reward distribution $p_t(r_t\mid s_t,a_t)$, for $t=1,\dots,H$. 
A policy $\pi=(\pi_t(a_t\mid s_t))_{t\leq H}$ induces a distribution over trajectories $\traj=(s_1,a_1,r_1,\dots,s_T,a_H,r_H)$:
\begin{align} \label{eq:trajdist_mdp}
p_\pi(\traj)=p_{1}(s_{1})\pi_1(a_1\mid s_1)p_1(r_1\mid s_1,a_1)\prod_{t=2}^{H}p_{t}(s_{t}\mid s_{t-1},a_{t-1})\pi_t(a_t\mid s_t)p_t(r_t\mid s_t,a_t).
\end{align}
In this paper, we focus on \emph{continuous} actions, $\Acal_t\subseteq\Rl$. For brevity we focus on the univariate case; the extension to multivariate actions is straightforward.
We are interested in the value, $J=\E_{p_{\epol}}[{\sum_{t=1}^Hr_t}]$, of a given policy, $\epol$, called the evaluation policy.
In particular, we consider the case where $\epol$ is \emph{deterministic} in that it is given by maps $\tau=(\tau_t)_{t\leq H},\,\tau_t:\mathcal S_t\to\mathbb R$ such that $\epol_t(a_t\mid s_t)=\delta(a_t-\tau_t(s_t))$ is the Dirac measure at $\tau_t(s_t)$, meaning when we follow $\epol$, $a_t$ is a function $\tau_t$ of $s_t$. When $\tau$ is parametrized as $\tau=\tau_\theta$ by some parameter $\theta\in\Theta$, we are also often interested in the policy gradient, $Z=\nabla_\theta J$, as it can be used for policy \emph{learning} via gradient ascent. We often drop the subscript and understand $\nabla$ to be with respect to (wrt ) $\theta$. When studying policy gradient estimation, we will assume throughout that $\tau_{\theta,t}(s_t)$ is almost surely differentiable in $\theta$ and that $\|\nabla\tau_t(s_t)\|_{\oper}\leq\Upsilon$ for some $\Upsilon<\infty$, where $\|\cdot\|_{\oper}$ is the matrix operator norm. Additionally, in theoretical results, we will assume actions are bounded: $\operatorname{support}(\bpol_t(\cdot\mid s_t))=[0,1]$, $\tau_t(s_t)\in(0,1)$.

In the \emph{offline} setting, the data available to us for estimating $J$ and $Z$ consists only of trajectory observations from some \emph{different} fixed policy, $\pi^b$, called the \emph{behavior policy}:
\begin{equation}\tag{off-policy data}\label{eq:offpolicydata}
\traj^{\braangle{1}},\dots,\traj^{\braangle{n}}\sim p_{\pi^b},\,\traj^{\braangle{i}}=(S^{\braangle{i}}_1,A^{\braangle{i}}_1,R^{\braangle{i}}_1,\cdots,S^{\braangle{i}}_H,A^{\braangle{i}}_H,R^{\braangle{i}}_H).
\end{equation}
Let $w^{\epol}_t(s_t)=p_{\epol}(s_t)/p_{\bpol}(s_t)$ denote the marginal density ratio, where $p_{\epol}(s_t),\,p_{\bpol}(s_t)$ are the marginal densities of $s_t$ under $p_{\epol}(\traj),\,p_{\bpol}(\traj)$, respectively. Let $q^{\epol}_t(s_t,a_t)=\E_{p_{\epol}}[{\sum_{k=t}^Hr_t\mid s_t,a_t}]$, $v^{\epol}_t(s_t)=\E_{p_{\epol}}[{\sum_{k=t}^Hr_t\mid s_t}]$ be $\epol$'s $q$- and $v$-functions.
We assume throughout that $1/\bpol_t(a_t\mid s_t)\leq C_1,\,w^{\epol}_t(s_t)\leq C_2$.
For clarity, we reserve capital letters for observed data, dropping superscripts $^{\braangle{i}}$ for a generic data point, and use lower case for generic MDP random variables. All expectations without subscripts are taken wrt $p_\bpol$. The \emph{empirical expectation} is $\ts\P_nf=\frac1n\sum_{i=1}^nf(\traj^{\braangle{i}})$. Define the $L_2$ norm as $\|f(\traj)\|_2=\E[f(\traj^2)]^{1/2}$. We let $\cdot^{\branormal{i}}$ denote the $i$-th order derivative wrt action alone and define the max Sobolev norm as $\|f\|_{j,\infty}=\max_{i=0,\cdots,j}\|f^{\branormal{j}} \|_{\infty}$. We emphasize that, e.g., $q^{(i)}(s_t,a_t)$ refers to differentiating only wrt $a_t$ alone.

\paragraph{Integral kernels}
In developing estimators we will use a second-order kernel $k:\Rl\to\Rl$, i.e., $\int k(u)\rD u=1$, $\int u k(u)\rD u=0$, $M_2(k)=\int u^2k(u)\rD u<\infty$.
We define $\Omega^{(i)}_2(k)=\int (k^{(i)}(u))^2\rD u$.
Given a bandwidth $h$, let $K_h(u)=h^{-1}k(u/h)$. 
Examples of differentiable kernels include 
Gaussian $k(u)\propto\exp(-u^2)$, biweight $k(u)\propto \max(0,1-u^2)^2$, and triweight $k(u)\propto \max(0,1-u^2)^3$.

\paragraph{Background on Offline Evaluation and Policy Gradient}

Direct estimation of $q$-functions \citep[direct method, DM;][]{munos2008finite} and step-wise importance sampling \citep[IS;][]{precup2000eligibility} are two common approaches for OPE. However, the former is known to suffer from the high variance and the latter from model misspecification. The doubly robust (DR) estimate combines the two, but its asymptotic MSE can still grow exponentially in horizon \citep{jiang,thomas2016}. \citet{KallusUehara2019} show that the efficient MSE in the MDP case is polynomial in $\bigO(H^2/n)$ and give an estimator achieving it by combining marginalized IS \citep{XieTengyang2019OOEf} and $q$-modeling using cross-fold estimation \citep{ChernozhukovVictor2018Dmlf}. OPE and off-policy policy gradient estimation are closely related \citep{HuangJiawei2019FISt}. Efficiency analysis and efficient estimators for off-policy policy gradients was recently given in \citet{EEOPG2020}.

All of the aforementioned methods assume that the density ratio $\epol_t(a_t\mid s_t)/\bpol_t(a_t\mid s_t)$ exists and is bounded. However, when $\epol$ is deterministic this is generally violated as it requires that $\bpol_t(a_t\mid s_t)$ has an atom at $\tau_t(s_t)$ but $\bpol$ usually has no atoms and at most can only have countably-many, while $\tau_t(s_t)$ can vary continuously, especially for policy learning (e.g., via gradient ascent).
In the bandit setting ($H=1$), \citet{pmlr-v84-kallus18a} recently showed that OPE is feasible under additional smoothness assumptions on induced action densities. The core idea is to approximate the deterministic policy by stochastic policy based on kernels. Namely, under appropriate smoothness,
\begin{align}\label{eq:ipw} 
    \lim_{h\to 0}\E\left[\frac{K_h(A_1-\tau_1(S_1))R_1}{\bpol(A_1\mid S_1)}\right]=\E_{\epol}\left[r_1\right].
\end{align}
\citet{pmlr-v84-kallus18a}, assuming known behavior policy, therefore propose an IS-type estimator $\P_n[\frac{K_h(A_1-\tau_1(S_1))R_1}{\bpol(A_1 \mid S_1)}]$ for $h$ appropriately shrinking in $n$ and analyze its bias and variance. %
This essentially amounts to approximating the deterministic policy with a stochastic one concentrated near, but not fully at, $\tau_1(S_1)$. Several works similarly deal with causal inference with continuous treatments \citep{ImaiKosuke2004CIWG,2005TPSw,GalvaoAntonioF2015USEE,FongChristian2018Cbps,WuXiao2018MoGP,KennedyEdwardH.2017Nmfd,SuLiangjun2019Nmwh} focusing on a single ($H=1$) constant ($\tau_1(s_1)=a_1^*$) action.
Although we do not explicitly use counterfactual notation, our estimand is equivalent to a counterfactual one under sequential ignorability \citep{hernan2019}. 

\section{Bandit Setting, \texorpdfstring{$H=1$}{a}}
For lucid presentation, we first develop off-policy value and gradient estimators in the bandit setting, i.e., $H=1$. In this section, since $H=1$, we omit the time index, e.g., letting $S=S_1$, $q=q_1$, etc.

\subsection{Off-Policy Deterministic Policy Evaluation}

There are actually several different ways to kernelize the deterministic policy. To motivate our estimators, note that, under appropriate smoothness (see \cref{thm:ope_final} below) we have that
\begin{align}\label{eq:dr} 
    \lim_{h\to 0}\E\left[\frac{K_h(A-\tau(S))\{R-f_{1}\}}{f_2}+f_3\right]=\E_{\epol}\left[r_1\right],
\end{align}
when each of $f_1,f_2,f_3$ takes any of the following two values, giving $2^3$ possible combinations:
\begin{align*}
f_1= \begin{cases} q(S,A) \\
    q(S,\tau(S))
    \end{cases},~f_2= \begin{cases} \bpol(A\mid S) \\
    \bpol(\tau(S)\mid S)
    \end{cases},~f_3= \begin{cases} 
    \int q(S,a)K_h(a-\tau(S))\rD a\\
    q(S,\tau(S))
    \end{cases}. 
\end{align*}
This general form includes the form of some previously proposed estimators,
including the IS-type estimator in \citet{pmlr-v84-kallus18a} ($f_1=0,f_2=\bpol(A\mid S),f_3=0$), the DR-type estimator in \citet{pmlr-v84-kallus18a} ($f_1=q(S,A),\,f_2=\bpol(A\mid S),\,f_3=q(S,\tau(S))$), and the DR-type estimator in \citet{Kyle2019} ($f_1=q(S,\tau(S)),\,f_2=\bpol(\tau(S)\mid S),\,f_3=q(S,\tau(S))$).

Based on this, we propose a general DR-type estimator with nuisance functions $f_1,f_2,f_3$: first, we split the data randomly into two halves $\cu_1$ and $\cu_2$; then, we define the estimator as 
\begin{align}
\label{eq:ope_estimators}
    \frac12\P_{\cu_1}\left[ \frac{K_h(A-\tau(S))\{R-\hat f^{[1]}_{1}\}}{\hat f^{[1]}_2}+\hat f^{[1]}_3\right]+\frac12\P_{\cu_2}\left[ \frac{K_h(A-\tau(S))\{R-\hat f^{[2]}_{1}\}}{\hat f^{[2]}_2}+\hat f^{[2]}_3\right], 
\end{align}
where $\hat f^{[k]}_j$ is an estimate of $f_j$ based on estimating $\pi^b,q$ by $\hat\pi^{b,[k]},\hat q^{[k]}$ using only the data in $\cu_{3-k}$.
This technique is called a cross-fitting, which is used to avoid metric entropy conditions on nuisance estimators \citep{ChernozhukovVictor2018Dmlf}.
All of our nuisances can be estimated by standard nonparametric density or regression estimators \citep{h09}. 
To simplify notation, we often write this and similar estimators as 
$\ts
\hat J=\P_n\bracks{{K_h(A-\tau(S))\{R-\hat f_{1}\}}/{\hat f_2}+\hat f_3}, 
$ where implicitly $\hat f_j$ are fit using cross-fitting on the half of the data that excludes the data point on which it is evaluated. 

We next analyze two primary cases: 
$f_1=q(S,A),\,f_2=\bpol(A\mid S),\,f_3=\int q(S,a)K_h(a-\tau(S))\rD a$, which refer to as $\hat J^\Kcal$ (for ``kernel''), and
$f_1=q(S,\tau(S)),\,f_2=\bpol(\tau(S)\mid S),\,f_3=q(S,\tau(S))$, which refer to as $\hat J^\Dcal$ (for ``deterministic'' as we plug in the deterministic policy into the nuisances).

\begin{theorem}
\label{thm:ope_final}
Suppose for $i=1,2$,
$\E[\|\hat \pi^{b,[i]}- \pi^b\|_{1,\infty}]=\smallo(1)$, $\E[\|\hat q^{[i]}- q\|_{1,\infty}]=\smallo(1)$, $\E[\|\hat \pi^{b,[i]}- \pi^b\|_{\infty}\|\hat q^{[i]}- q\|_{\infty}]=\smallo((nh)^{-1/2})$, $nh^5=\bigO(1)$, $nh\to \infty$, that $\bpol(a\mid s),\,q(s,a)$ are twice continuously differentiable wrt $a$ for almost all $s$, and that $\hat \pi^{b,[i]},\hat q^{[i]}$ are uniformly bounded by a constant.
Then, the bias and variance of $\hat J^{\Dcal}$ are $\E[\hat J^{\Dcal}]-J=0.5M_2(k)h^2B+\smallo((nh)^{-1/2}),\,\var[\hat J^{\Dcal}]=\frac{\Omega_2(k)}{nh}(V+\smallo(1))$, where
\begin{align*}
  B=\E[q^{(2)}(S,\tau(S))+2q^{(1)}(S,\tau(S))\pi^{b(1)}(\tau(S)|S)/\pi^{b}(\tau(S)|S)],\quad V=\E\left[\frac{\mathrm{var}[R\mid S,\tau(S)]}{\pi^b(\tau(S)\mid S) } \right]. 
\end{align*}
If additionally $\hat\pi^{b,[i]}(a \mid s),\,\hat q^{[i]}(s,a)$ are twice continuously differentiable wrt $a$, then the same holds for $\hat J^{\Kcal}$ with $B=\E[q^{\branormal{2}}(S,\tau(S))]$ and the same $V$ as the above. In both cases, setting $h=\Theta(n^{-1/5})$ yields the minimal MSE of order $\bigO(n^{-4/5})$.
\end{theorem}
Here, we assume that nuisance estimation errors converge in expectation (i.e., in $L_1$). If we instead assume the weaker convergence in probability, we can obtain the same guarantee on the bias and variance, conditioned on an event that occurs with high probability. Refer to \cref{sec:probability}. 

\begin{remark}
Three things should be noted. 
First, the best MSE rate achievable in the result is $\bigO(n^{-4/5})$%
, which is slower than the usual rate $\bigO(n^{-1})$ in OPE of stochastic policies under positivity. This slow rate is expected because our estimand in the deterministic case is not regular \citep{KennedyEdwardH.2017Nmfd}. Since the minimax rate for density estimation in a Sobolev class of smoothness parameter $2$ is $\bigO(n^{-4/5})$ \citep{KorostelevAlexander}, we expect that this rate is minimax for $J$ among problems satisfying the conditions of \cref{thm:ope_final}. Establishing this formally is future work. Second, for $\hat J^\Dcal$, the only condition on nuisance estimators is a sub-parametric rate, which can, e.g., be satisfied when each nuisance converges at the rate $\smallo(n^{-1/5})$ for $h=\Theta(n^{-1/5})$. This condition appears weaker than the $\smallo(n^{-1/4})$ nuisance rate required in usual OPE \citep{KallusUehara2019,ChernozhukovVictor2018Dmlf}; however, the required norm itself is stronger since $\|\cdot\|_2\leq \|\cdot\|_{\infty}$. In contrast, when just $q$ is estimated at rate $\smallo(n^{-1/5})$, one cannot guarantee a similar $\bigO(n^{-4/5})$ MSE rate for DM without additional assumptions; a simple triangle inequality yields only an $\smallo(n^{-2/5})$ MSE rate. The same is true for IS when we estimate $\bpol$ only at $\smallo(n^{-1/5})$ rate.\footnote{In the special case when $q$ (or $\pi^b$, respectively) is estimated using kernel estimators, we can obtain an MSE rate $\bigO(n^{-4/5})$ for DM (or IS, respectively) but it requires much stricter conditions on the smoothness, including smoothness in the state variable in addition to smoothness in action \citep{HsuYu-Chin2018Daie,Ying-YingLee2018PMPw}.} 
Third, both the variances and rates for $\hat J^\Dcal$ and $\hat J^\Kcal$ are the same, while the bias constant is slightly different.
For brevity, in the following, we focus on deriving theoretical properties for $\hat J^\Kcal$ where similar results are easily obtainable for $\hat J^\Dcal$.
\end{remark}

\paragraph{Optimality of $\hat J^{\Kcal}$ in terms of leading constant}
\citet{pmlr-v84-kallus18a} computed the asymptotic bias and variance for the IS estimator ($f_1=0,f_2=\bpol(A\mid S),f_3=0$) with known behavior policy (hence no nuisances). The estimator $\hat J^{\Kcal}$ is obtained by adding the control variate $f_1=f_3=q(S,A)$ and while the two estimators have the \emph{same} asymptotic bias, the leading variance term is \emph{smaller}, having $\E\left[\frac{\mathrm{var}[R\mid S,\tau(S)]}{\pi^b(\tau(S)\mid S) } \right]$ instead of the larger $\E\left[\frac{\mathrm{\E}[R^2 \mid S,\tau(S)]}{\pi^b(\tau(S)\mid S) } \right]$ in \citet{pmlr-v84-kallus18a}.
Thus, the advantage of our DR-type estimators over the IS estimator is not only ensuring $\bigO(n^{-4/5})$ convergence with an \emph{unknown} behavior policy but also in providing an improvement in the variance leading term. In fact, we can prove $\hat J^{\Kcal}$ is optimal among a class of estimators. 
\begin{corollary}\label{cor:optimal}
Assume $\bpol(a|s),f(s,a),q(s,a)$ are $C^2$-functions wrt $a$. 
Then $\P_n[\frac{K_h(A-\tau(S))}{\bpol(A|S)}\{R-f(S,A)\}+f(S,A)]$ has a bias independent of $f$ and variance minimized by letting $f=q$.
\end{corollary}
This would correspond to an efficiency result in semiparametric theory, but that cannot applied here since our estimand is not regular \citep{KennedyEdwardH.2017Nmfd}. It is also difficult to compare $\hat J^{\Kcal}$ and $\hat J^{\Dcal}$ since the bias terms are different. We leave further investigation of optimality to future work.

\begin{remark}[Relation with previous literature]

 \citet{BibautAurelien2017Dsfo} proposed a general approach for the estimation of non-regular estimands using smoothing. See the estimator in Example 3; however, they assumed a behavior policy is known. \citet[Section 8]{FosterDylanJ.2019OSL} also touched on the idea of case $\Kcal$. However, they did not analyze a mathematical detail. 
\end{remark}

\subsection{Off-Policy Deterministic Policy Gradient Estimation}

For a deterministic policy class $\{\tau_{\theta}(s):\theta \in \Theta\}$, consider estimating
$Z=\nabla J$ at a given $\theta$.
Usually policy gradients involve the policy score, $\nabla \log(\epol_\theta(a\mid s))$ \citep{PetersJ2006PGMf,EEOPG2020}. However, for deterministic policies, these policy scores do not exist. However, assuming that $q$ is differentiable in $a$ immediately yields $Z=\E[q^{(1)}(S,\tau_{\theta}(S)) \nabla \tau_{\theta}(S)]$, suggesting this may still be possible under appropriate smoothness.

\paragraph{Deterministic Policy Gradient (DPG) and IS Policy Gradient (ISPG)}

By taking a derivative of the IS and DM estimators wrt $\theta$, we obtain corresponding policy gradient estimators: 
\begin{align}\label{eq:dpg}
\hat Z^\text{IS}=\P_n[K^{(1)}_h(A-\tau(S))R/\hat \pi^b(A\mid S)],\quad \hat Z^\text{DPG}=\P_n[\hat q^{(1)}(S,\tau_{\theta}(S))\nabla_{\theta}\tau_{\theta}(S)],
\end{align}
where  $K^{(1)}_h(u)=-h^{-2}k^{(1)}(u/h)$. The latter estimator is a bandit version of DPG \citep{silver14}. Like their OPE counterparts, these estimators suffer from high dependence on the nuisance estimates and potentially slow rates.

\paragraph{Doubly Robust DPG}

By differentiating our OPE estimator $\hat J^{\Kcal}$ wrt $\theta$ 
we propose a new policy gradient estimator: $\hat Z^{\Kcal}=\P_n[\psi^{\Kcal}(\hat q,\hat \pi^{b})]$ (recall $\hat q,\hat\pi^b$ are implicitly cross-fit in this notation), where
\begin{align}
\label{eq:pg2_bandit}
\psi^{\Kcal}(q,\pi^b)= \left\{\frac{K^{(1)}_h(a-\tau_{\theta}(s))\{r-q(s,a) \}}{\bpol(a\mid s)}+\int K^{(1)}_h(a-\tau_{\theta}(s))q(s,a)\rD a \right\}\nabla_{\theta}\tau_{\theta}(s).
\end{align}

Notice this does not involve explicit differentiation of $q$. Instead the convolution with $K^{(1)}_h$ essentially acts as an estimator for the derivative and may be computationally more stable than differentiating $\hat q$.

Similarly, by differentiating our OPE estimator  $\hat J^{\Dcal}$ wrt $\theta$,
we propose $\hat Z^{\Dcal}=\P_n[\psi^{\Dcal}(\hat q,\hat \pi^{b})]$, where
\begin{align}
\label{eq:pg3_bandit}
\psi^{\Dcal}(q,\pi^b)= \left\{\frac{K^{(1)}_h(a-\tau_{\theta}(s))\{r-q(s,\tau_{\theta}(s)) \}}{\bpol(\tau_{\theta}(s) \mid s)}+ q^{(1)}(s,\tau_{\theta}(s))\right\}\nabla_{\theta}\tau_{\theta}(s). 
\end{align} 
This can be understood as DPG \emph{plus} a control variate. As in DPG, $\hat Z^{\Dcal}$ requires differentiation of $\hat q$.

\begin{theorem}\label{thm:theory_pg}
Suppose for $i=1,2$,
$\E[\|\hat \pi^{b,[i]}-\pi^b\|_{2,\infty}]=\smallo(1)$, $\E[\|\hat q^{[i]}-q\|_{2,\infty}]=\smallo(1)$, $\E[\|\hat \pi^{b,[i]}-\pi^b\|_{1,\infty}\|\hat q^{[i]}-q\|_{1,\infty}]=\smallo(n^{-1/2}h^{-3/2})$, $nh^{7}=\bigO(1)$, $nh\to\infty$, $q(s,a),\bpol(a\mid s),\hat q^{[i]}(s,a),\hat \pi^{b,[i]}(a\mid s)$ are thrice continuously differentiable functions wrt $a$, and $\hat q^{[i]},\hat \pi^{b,[i]}$ are uniformly bounded by a constant. Then, the bias and variance of $\hat Z^{\Kcal}$ are  $\E[\hat Z^{\Kcal}]-Z=0.5h^2M_2(k)\tilde{B}+\smallo(n^{-1/2}h^{-3/2}),\, \var[\hat Z^{\Kcal}]=\frac{\Omega^{(1)}_2(k)}{nh^3}\{\tilde{V}+\smallo(1)\}$, where 
\begin{align*}
\tilde{B}=\E\left[\nabla \tau_{\theta}(S)q^{(3)}(S,\tau_{\theta}(S))\right] ,\quad \tilde{V}=\E\left[\otimes \nabla \tau_{\theta}(S)\frac{\mathrm{var}[R\mid S,\tau_{\theta}(S)]}{\bpol(\tau_{\theta}(S) \mid S)}\right],
\end{align*}
where $\otimes v=vv^{\top}$.
Setting $h=\Theta(n^{-1/7})$ yields the minimal MSE of order $\bigO(n^{-4/7})$. 
\end{theorem}
\begin{remark}
We can obtain a similar result for $\hat Z^{\Dcal}$ with slightly different differentiability conditions; we omit the details.
The best-achievable MSE rate in \cref{thm:theory_pg}, $\bigO(n^{-4/7})$, matches the minimax rate for density gradient estimation in a Sobolev space of smoothness parameter $3$ \citep{KorostelevAlexander}. We therefore conjecture the rate for $Z$ is minimax optimal under the assumptions of the theorem.
\end{remark}
\begin{remark}
As in the case of OPE,
there are two crucial advantages of $\hat Z^{\Kcal},\hat Z^{\Dcal}$ over ISPG and DPG estimators in \cref{eq:dpg}. First, the required convergence rates on nuisances are weaker and we do not depend on the particular estimators. On the other hand, ISPG and DPG do not have convergence guarantees given only rate conditions on $\hat\pi^b$ and $\hat q$, respectively. Second, our leading constant in the variance is smaller than ISDP with an oracle behavior policy, just as in \cref{cor:optimal}.
\end{remark}
\begin{remark}[$\hat Z^\Kcal$ vs $\hat Z^\Dcal$]\label{rem:comparison}
Unlike $\hat Z^{\Kcal}$,
the estimator $\hat Z^{\Dcal}$ does not involve integration, which may be computationally preferable. However, direct differentiation of $q$-functions can often be statistically unstable \citep[see also][Section 5.2]{AtheySusan2017EPL}. In our empirical results in \cref{sec:experiments}, we indeed find $\hat Z^{\Kcal}$ is superior. Moreover, when we use the Gaussian kernel and polynomial sieve nuisance estimators, the integration can be easily done analytically.
\end{remark}

\section{Offline RL with Deterministic Policies}

We next discuss how to extend the ideas from the previous section to the RL setting where $H\geq1$ in general. In this setting there are actually different ways to account for the IS part of the estimator, leading to different dependence on horizon.
Throughout this section, we will assume the densities $p_t(r_t \mid  s_t,a_t),p_{t+1}(s_{t+1}\mid s_t,a_t)$ are thrice continuously differentiable wrt action and  $\|\int |p^{(i)}_t(r_t\mid s_t,a_t)|\rD r_t\|_{\infty}\leq G^{(i)}_1<\infty,\,\|\int| p^{(i)}_{t}(s_t\mid s_{t-1},a_{t-1})|\rD s_t\|_{\infty}\leq G^{(i)}_2<\infty$ for $ i=1,2,3$.  We also assume all of nuisance estimators introduced in this section are uniformly bounded by some constant. 
\subsection{Off-Policy Deterministic Policy Evaluation}\label{sec:ope_rl}

Motivated by DR OPE using cumulative density ratios for the case of stochastic policies \citep{jiang}, we propose analogous extensions of $\hat J^{\Kcal},\hat J^{\Dcal}$ for $H\geq1$: 
Cumulative DR case $\Kcal$ (CDRK)
$\hat J^{\Kcal}_{\CD}=\P_n[\phi^{\Kcal}_{\CD}(\hat q^{\Kcal}_t,\hat \pi^b_t)]$ and Cumulative DR case $\Dcal$ (CDRD) $\hat J^{\Dcal}_{\CK}=\P_n[\phi^{\Dcal}_{\CK}(\hat q^{\epol}_t,\hat \pi^b_t)]$, where
\begin{align*}
   \phi_\CD^{\Kcal}(q^{\Kcal}_t,\bpol_t)&=v^{\Kcal}_1+\sum_{t=1}^H \lambda^{\Kcal}_t\{r_t-q^{\Kcal}_t(s_t,a_t)+v^{\Kcal}_{t+1}\},\,
  \lambda^{\Kcal}_t=\prod_{k=1}^t \frac{K_h(a_k-\tau_k(s_k))}{\bpol_k(a_k\mid s_k)},
   \\
   \phi_\CK^{\Dcal}(q^{\epol}_t,\bpol_t)&=v^{\epol}_1+\sum_{t=1}^H \lambda^{\epol}_t\{r_t-q^{\epol}_t(s_t,\tau_t(s_t))+v^{\epol}_{t+1}\},\,
   \lambda^{\epol}_t=\prod_{k=1}^t \frac{K_h(a_k-\tau_k(s_k))}{\bpol_k(\tau_k(s_k)\mid s_k)},
\end{align*}
and where $q_t^\Kcal$ is the $q$-function associated with the kernelized evaluation policy, $\pi^{e,\Kcal}_t(a_t\mid s_t)=K_h(a_t-\tau_{\theta}(s_t))$.
We discuss the estimation of nuisances in \cref{rem:nuisance}. Recall we use cross-fitting.

\begin{theorem}\label{thm:rl_ope}
\label{thm:OPE_RL}
Suppose for $j\leq H,i=1,2$, 
$\E[\|\hat \pi^{b,[i]}_j- \bpol_j\|_{1,\infty}]=\smallo(1)$,
$\E[\|\hat q^{\Kcal,[i]}_j-q^{\Kcal}_j\|_{1,\infty}]=\smallo(1)$,
$\E[\|\hat \pi^{b,[i]}_j- \bpol_j\|_{\infty}\|\hat q^{\Kcal,[i]}_j-q^{\Kcal}_j\|_{\infty}]=\smallo(n^{-1/2}h^{-H/2})$,
$nh^{H+4}=\bigO(1)$, $nh\to\infty$. Then, we have $\E[\hat J^{\Kcal}_{\CK}]-J=0.5h^2M_{2}(k)B^{\CD}_H+\smallo(n^{-1/2}h^{-H/2})$, $\var[\hat J^{\Kcal}_{\CK}]=\frac{\Omega^H_2(k)}{nh^{H}} \{V^{\CD}_H+\smallo(1)\}$, where 
\begin{align*}
    B^{\CD}_H & = \sum_{t=1}^H\E_{p_\epol}\left[\frac{r_tp^{(2)}_t(r_t\mid s_t,\tau_t(s_t)) }{p_t(r_t\mid s_t,\tau_t(s_t))} \right] +\sum_{j=1}^{t-1}\E_{p_\epol}\left[\frac{r_tp^{(2)}_{j+1}(s_{j+1}\mid s_j,\tau_j(s_j))}{p_{j+1}(s_{j+1}\mid s_j,\tau_j(s_j))}\right ],  \\
 V^{\CD}_H &= \E_{p_\epol}\left[\frac{1}{\prod_{i=1}^{H}\pi^b_i(\tau_i(s_i)\mid s_i))}\mathrm{var}[r_H\mid  s_H,a_H]\right].  
\end{align*}
In the above, setting $h=\Theta(n^{-1/(H+4)})$ yields the minimal MSE of order $\bigO(n^{-4/(H+4)})$.
\end{theorem}

\begin{remark}[Curse of Horizon in Rate]
Notice CDRK and CDRD have convergence \emph{rate} that deteriorates as the horizon grows.
In usual OPE for stochastic policies, \citet{Liu2018,KallusNathan2019EBtC,KallusUehara2019} show that using cumulative IS leads to MSE with leading constant that grows exponentially in horizon, but it still has rate $\bigO(1/n)$.
Thus, the curse of horizon is even more detrimental for deterministic policies. However, CDRK and CDRD technically work also for \emph{non}-Markov decision processes. Next, we will tackle the curse in rate by leveraging Markovian structure, following \citet{KallusUehara2019}.
\end{remark}

Motivated by DR OPE using marginal density ratios for stochastic policies \citep{KallusUehara2019}, we propose 
Marginal DR case $\Kcal$ (MDRK) for deterministic OPE: $\hat J^{\Kcal}_{\mathrm{mdr}}=\P_n[\phi^{\Kcal}_{\mathrm{mdr}}(\hat q^{\Kcal}_t,\hat w^{\Kcal}_t,\hat \pi^b_t) ]$,
where 
\begin{align}
 \phi_{\mathrm{mdr}}^{\Kcal}(q^{\Kcal}_t,w^{\Kcal}_t,\bpol_t)=v^{\Kcal}_1(s_1)+\sum_{t=1}^{H} {w^{\Kcal}_t(s_t)}{\bpol_t(a_t\mid s_t)}K_h\left(a_t-\tau_t(s_t)\right)(r_t-q^{\Kcal}_t(s_t,a_t)+v^{\Kcal}_{t+1}(s_{t+1})) 
 \notag
\end{align}
and $w^{\Kcal}_t(s_t)=p_{\pi^{e,\Kcal}}(s_t)/p_{\bpol}(s_t)$ is the marginal density ratio associated with $\pi^{e,\Kcal}_t$.
Again, a similar estimator, Marginal DR case $\Dcal$ (MDRD), $\hat J^{\Dcal}_{\mathrm{mdr}}$, is constructed by replacing $q^{\Kcal}_t(s_t,a_t),w^{\Kcal}_t(s_t),\bpol_t(a_t\mid s_t)$ with $q^{\epol}_t(s_t,\tau_t(s_t)),w^{\epol}_t(s_t),\bpol_t(\tau_t(s_t)\mid s_t)$. 
\begin{theorem}\label{thm:rl_ope2}
Suppose for $j\leq H,i=1,2$, 
$\E[\|\hat \pi^{b,[i]}_j- \bpol_j\|_{1,\infty}]=\smallo(1)$,
$\E[\|\hat w^{\Kcal,[i]}_j- w^\Kcal_j\|_{\infty}]=\smallo(1)$,
$\E[\|\hat q^{\Kcal,[i]}_j-q^{\Kcal}_j\|_{1,\infty}]=\smallo(1)$,
$\E[\max\{\|\hat \pi^{b,[i]}_j- \pi^b_j\|_{\infty},\|\hat w^{\Kcal,[i]}_j- w^{\Kcal}_j\|_{\infty}\}\|\hat q^{\Kcal,[i]}_j-q^{\Kcal}_j\|_{\infty}]=\smallo(n^{-1/2}h^{-1/2})$,
$nh^{5}=\bigO(1)$, $nh\to\infty$.
Then, the bias of $\hat J^{\Kcal}_{\mathrm{mdr}}$ is the same as $\hat J^{\Kcal}_{\mathrm{cdr}}$ in \cref{thm:OPE_RL} and its variance is
$\var[\hat J^{\Kcal}_{\mathrm{mdr}}]=\frac{\Omega_2(k)}{nh} V^{\mathrm{mdr}}_H+\smallo(n^{-1}h^{-1})$, where 
\begin{align*}
 V^{\mathrm{mdr}}_H=\sum_{t=1}^{H} \E_{p_\epol}\left[\frac{w^{\epol}_t(s_t)}{\bpol_t(\tau_t(s_t)\mid s_t)}\mathrm{var}[r_t+v^{\epol}_{t+1}(s_{t+1})\mid s_t,\tau_t(s_t) ] \right].
\end{align*}
Setting $h=\Theta(n^{-1/5})$ yields the minimal MSE of order $\bigO(n^{-4/5})$. Specifically, if $h=cn^{-1/5}$, then
\begin{align*}
\E[(\hat J^{\Kcal}_{\mathrm{mdr}}-J)^2]\leq
  n^{-4/5}R^2_{\max}H^2\braces{ \frac{c^{4}M^2_2(k)}{4}(G^{(2)}_1+\frac{(H-1)}{2}G^{(2)}_2)^2+\frac{C_1C_2\Omega_2(k)}{c}}+\smallo(n^{-4/5}).
\end{align*}
\end{theorem}
Notice the minimal MSE rate is the same as in the bandit case. We therefore conjecture the rate to be minimax optimal. More crucially, it does not suffer from the curse of horizon in rate.
Moreover, the dependence of the leading constant is polynomial in horizon, $\bigO(H^4)$.
The leading constant is smaller when $C_1,C_2,G^{(2)}_1,G^{(2)}_2$ are smaller, i.e., when the behavior policy is closer to the evaluation policy and the reward and transition densities are smoother.

\subsection{Off-Policy Deterministic Policy Gradient Estimation}

We next construct deterministic policy gradient estimators for RL.
By differentiating $\phi_{\mathrm{mdr}}^{\Kcal}$, we obtain the Marginal PG case $\Kcal$ (MPGK) estimator, $\hat Z^{\Kcal}_{\MK}=\P_n[\psi^{\Kcal}_{\MK}( \hat q^{\Kcal}_t,\hat w^{\Kcal}_t,\hat d^{q^{\Kcal}}_t,\hat d^{w^{\Kcal}}_t,\hat \pi^b_t)]$,
where 
\begin{align*} 
&\psi^{\Kcal}_\MK(q^{\Kcal}_t,w^{\Kcal}_t,d^{q^{\Kcal}}_t,d^{w^{\Kcal}}_t,\bpol_t)=
d^{v^{\Kcal}}_1+\sum_{t=1}^{H} \frac{d^{w^{\Kcal}}_tK_h\left(a_t-\tau_t(s_t)\right)}{\bpol_t(a_t|s_t)}(r_t-q^{\Kcal}_t+v^{\Kcal}_{t+1}) \nonumber \\
&~~~~+\sum_{t=1}^{H} \frac{{w_t^{\Kcal}}}{\bpol_t(a_t|s_t)}\Bigl(K^{(1)}_h\left(a_t-\tau_t(s_t)\right)(r_t-q^{\Kcal}_t+v^{\Kcal}_{t+1})\nabla \tau_t(s_t)+K_h\left(a_t-\tau_t(s_t)\right)(-d^{q^{\Kcal}}_t+d^{v^{\Kcal}}_{t+1})\Bigr),\\
&q^{\Kcal}_t=q^{\Kcal}_t(s_t,a_t), d^{q^{\Kcal}}_t(s_t,a_t)=\nabla q^{\Kcal}_t(s_t,a_t),\,d^{v^{\Kcal}}_t(s_t)=\nabla v^{\Kcal}_t(s_t),\,d^{w^{\Kcal}}_t(s_t)=\nabla w^{\Kcal}_t(s_t),\\
& v^{\Kcal}_t(s_t)=\int  q^{\Kcal}_t(s_t,a_t)K_h\left(a_t-\tau_t(s_t)\right)\rD a_t,\\& d^{v^{\Kcal}}_{t}(s_t)=\int \braces{ d^{q^{\Kcal}}_{t}(s_t,a_t)K_h\left(a_t-\tau_t(s_t)\right)+  q_t(s_t,a_t)K^{(1)}_h\left(a_t-\tau_t(s_t)\right) \nabla \tau_t(s_t)} \rD a_t.
\end{align*}
Notice we only estimate the nuisances $q^{\Kcal}_t,w^{\Kcal}_t,d^{q^{\Kcal}}_t,d^{w^{\Kcal}}_t,\bpol_t$; then estimates for $v^{\Kcal}_t,d^{v^{\Kcal}}_{t}$ are defined in terms of these.
The Marginal PG case $\Dcal$ (MPGD) estimator, $\hat Z^{\Dcal}_{\MD}$, is similarly defined 
by replacing the nuisances in $\hat Z^{\Kcal}_{\MD}$ by $q^{\epol}_t(s_t,\tau_t(s_t)), w^{\epol}_t(s_t), d^{q^{\epol}}_t(s_t,\tau_t(s_t)), d^{w^{\epol}}_t(s_t), \pi^b_t(\tau_t(s_t)\mid s_t)$.
Again, note $\hat Z^{\Dcal}_{\MD}$  involves a differentiation while $\hat Z^{\Kcal}_{\MD}$ does not but involves an integration, as in \cref{rem:comparison}.

\begin{theorem}\label{thm:gradient_est}
Suppose for $j\leq H,i=1,2$, 
$\E[\|\hat w^{\Kcal,[i]}_j-w^{\Kcal}_j\|_{\infty}]=\smallo(1)$,
$\E[\|\hat \pi^{b,[i]}-\pi^b\|_{2,\infty}]=\smallo(1)$,
$\E[\|\hat d^{w^{\Kcal},[i]}_j-d^{w^{\Kcal}}_j\|_{\infty}]=\smallo(1)$,
$\E[\|\hat q^{\Kcal,[i]}_j-q^{\Kcal}_j\|_{2,\infty}]=\smallo(1)$,
$\E[\|\hat d^{q^{\Kcal},[i]}_j-d^{q^{\Kcal}}_j\|_{1,\infty}]=\smallo(1)$,
$\E[\max\{\|\hat w^{\Kcal,[i]}_j-w^{\Kcal}_j\|_{\infty},\|\hat \pi^{b,[i]}_j-\pi^b_j\|_{1,\infty},\|\hat d^{w^{\Kcal},[i]}_j-d^{w^{\Kcal}}_j\|_{\infty}\}\max\{ \|\hat q^{\Kcal,[i]}_j-q^{\Kcal}_j\|_{1,\infty},\|\hat d^{q^{\Kcal},[i]}_j-d^{q^{\Kcal}}_j\|_{\infty}\}]=\smallo(n^{-1/2}h^{-3/2})$,
$nh^{7}=\bigO(1)$, $nh\to\infty$.
Then, we have $\E[\hat Z^{\Kcal}_{\MK}]-Z=0.5h^2M_2(k)\tilde B^{\mathrm{mpg}}_H+\smallo(n^{-1/2}h^{-3/2})$, $\var[\hat Z^{\Kcal}_{\MK}]=\frac{\Omega^{(1)}_2(k)}{nh^3}\tilde V^{\mathrm{mpg}}_H+\smallo(n^{-1}h^{-3})$, where 
\begin{align*}
\tilde B^{\mathrm{mpg}}_H  &=  \sum_{t=1}^{H}\left(\nabla \E_{p_\epol}\left[\frac{r_t p^{(2)}_t(r_t\mid s_t,\tau_t(s_t))}{p_t(r_t\mid s_t,\tau_t(s_t))} \right] +\sum_{l=1}^{t-1}\nabla \E_{p_\epol}\left[\frac{r_tp^{(2)}_{l+1}(s_{l+1}\mid s_l,\tau_l(s_l))}{p_{l+1}(s_{l+1}\mid s_l,\tau_l(s_l))}\right ] \right),\\
\tilde V^{\mathrm{mpg}}_H &= \sum_{t=1}^H \E_{p_\epol}\left[\frac{w^{\epol}_t(s_t) }{\pi^{b}_{t}(\tau_t(s_t)\mid s_t)}\var_{p_\epol}[r_t+q^{\epol}_{t+1}(s_{t+1},\tau_{t+1}(s_{t+1})) \mid s_t,\tau_t(s_t)]\otimes\nabla \tau_t(s_t)\right]. 
\end{align*}
Setting $h=\Theta(n^{-1/7})$ yields the minimal MSE of order $\bigO(n^{-4/7})$. Specifically, if $h=cn^{-1/7}$, the operator norm of the MSE is bounded by 
\begin{align*}\ts
\frac{R^2_{\max}H^2\Upsilon^2}{n^{4/7}}\bigl(\frac{c^4M^2_2(k)}{4}\braces{G^{(3)}_1+\frac{(H-1)}{2}\{G^{(2)}_1G^{(1)}_2+G^{(1)}_1G^{(2)}_2+G^{(3)}_2 \}+\frac{(H-1)(H-2)}{3} G^{(1)}_2G^{(2)}_2 }^2+\frac{C_1C_2\Omega^{(1)}_2(k)}{c^3}\bigr). 
\end{align*}
\end{theorem}
Note again that the MSE rate $\bigO(n^{-4/7})$ is slower than the usual $\bigO(1/n)$ efficient rate for off-policy gradient estimation with stochastic policies \citep{EEOPG2020}. Nonetheless, it matches the bandit case and we therefore conjecture it is minimax optimal. More importantly, we note that it alleviates the curse of horizon, both in rate and in leading constant.
We can also derive corresponding CPGK and CPGD estimators by differentiating the CDRK and CDRD estimating functions, but these \emph{will} suffer from the curse of horizon in rate, as in \cref{thm:OPE_RL}. 

\begin{remark}[Estimation of nuisance functions]\label{rem:nuisance}
Our OPE and off-policy policy gradient estimators depend on estimating some nuisances. A unique and new feature of our estimators and analysis compared to previous deterministic off-policy estimators is that the MSE guarantees do not depend on the particular nuisance estimator used and we make no assumptions except for their (slow) convergence rate. The estimation of $q_t,w_t$ for stochastic policies is discussed in \citet{KallusUehara2019}
and of $d^{q}_t,d^{w}_t$ in \citet{EEOPG2020}.
These can be applied directly to estimate $q^{\Kcal}_t,w^{\Kcal}_t,d^{q^{\Kcal}}_t,d^{w^{\Kcal}}_t$ since the kernelized evaluation policy, $\pi^{e,\Kcal}$, is stochastic.
The estimation of $q^{\epol}_t$ is the same for deterministic policies and a small adjustment can also be made for $d^{q^{\epol}}_t$ as we explain in \cref{sec:nuisance}. The estimation $w^{\epol}_t,d_t^{w^{\epol}}$ for deterministic policies is difficult, but we can simply use estimates of $w^{\Kcal}_t,d^{w^{\Kcal}}_t$ as estimates for $w^{\epol}_t,d_t^{w^{\epol}}$, which is essentially a kernel density estimation approach for the densities in the latter.
For additional detail, refer to \cref{sec:nuisance}. 
\end{remark}

\begin{remark}[Policy learning algorithms]
To do offline RL to learn a deterministic policy, we can combine
any type of gradient-based optimization algorithm with our estimated gradients. A simple gradient ascent is given as an example in \cref{sec:optimizaiton} and used in the experiments in the next section. Following \citet{EEOPG2020} we can also combine standard results for gradient ascent with our error bounds to obtain a regret guarantee. Since the proof is exactly the same, simply plugging in our error bounds instead, we omit the details and refer the reader to \citet{EEOPG2020}.
\end{remark}

\section{Experiments} \label{sec:experiments}

We next conduct an experiment in a very simple environment to confirm the theoretical guarantees of the proposed estimators. More extensive experimentation remains future work. The setting is as follows. Set $\mathcal{S}_t=\Rl,\,\mathcal{A}_t=\Rl,\,s_0=0$. Then, set the transition dynamics as $s_t=a_{t-1}-s_{t-1}+\mathcal{N}(0,0.3^2)$, the reward as $r_{t}=-s_t^2$, the behavior policy as $\bpol(a\mid s)=\mathcal{N}(0.8s,1.0^2)$, the deterministic evaluation policy as $\tau_t(s_t)=\theta s_t$, and the horizon as $H=20$. Note that in this setting, the optimal policy is given by $\theta^*=1$ . We compare CPGK, CPGD, MPGK, MPGD using the Gaussian kernel with PG. The nuisance functions $q,\,w,\,d^{q},\,d^{w}$ (and their case $\Kcal$ equivalents) are estimated using polynomial sieve regressions \citep{ChenXiaohong2007C7LS}. We assume the behavior policy is known. 
Since $q$ is estimated by polynomials and $k$ is Gaussian, we can compute the integrals in MPGK and CPGK analytically. 
We use the same estimated $q$ in PG. We choose $h$ by bootstrapping the estimator for each of $h\in\{0.05,0.1,0.25,0.5\}$ and choosing that with smallest bootstrap variance.

\begin{figure}
\centering
\begin{minipage}{.47\textwidth}
  \centering
  \includegraphics[width=.85\linewidth]{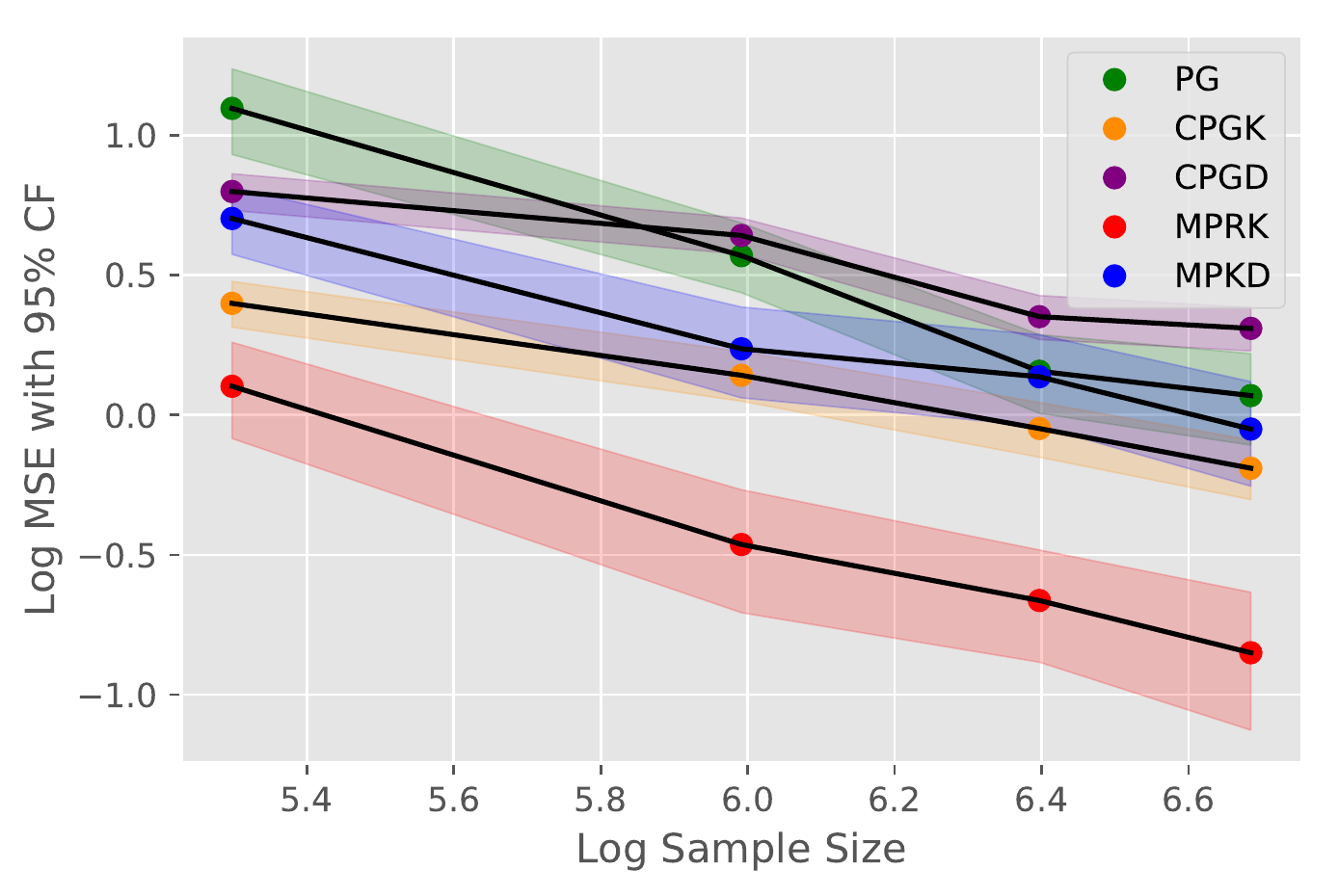}
  \captionof{figure}{MSE of gradient estimation with 95\% CI}
  \label{fig:gradient} 
\end{minipage}\hfill%
\begin{minipage}{.47\textwidth}
  \centering
  \includegraphics[width=.85\linewidth]{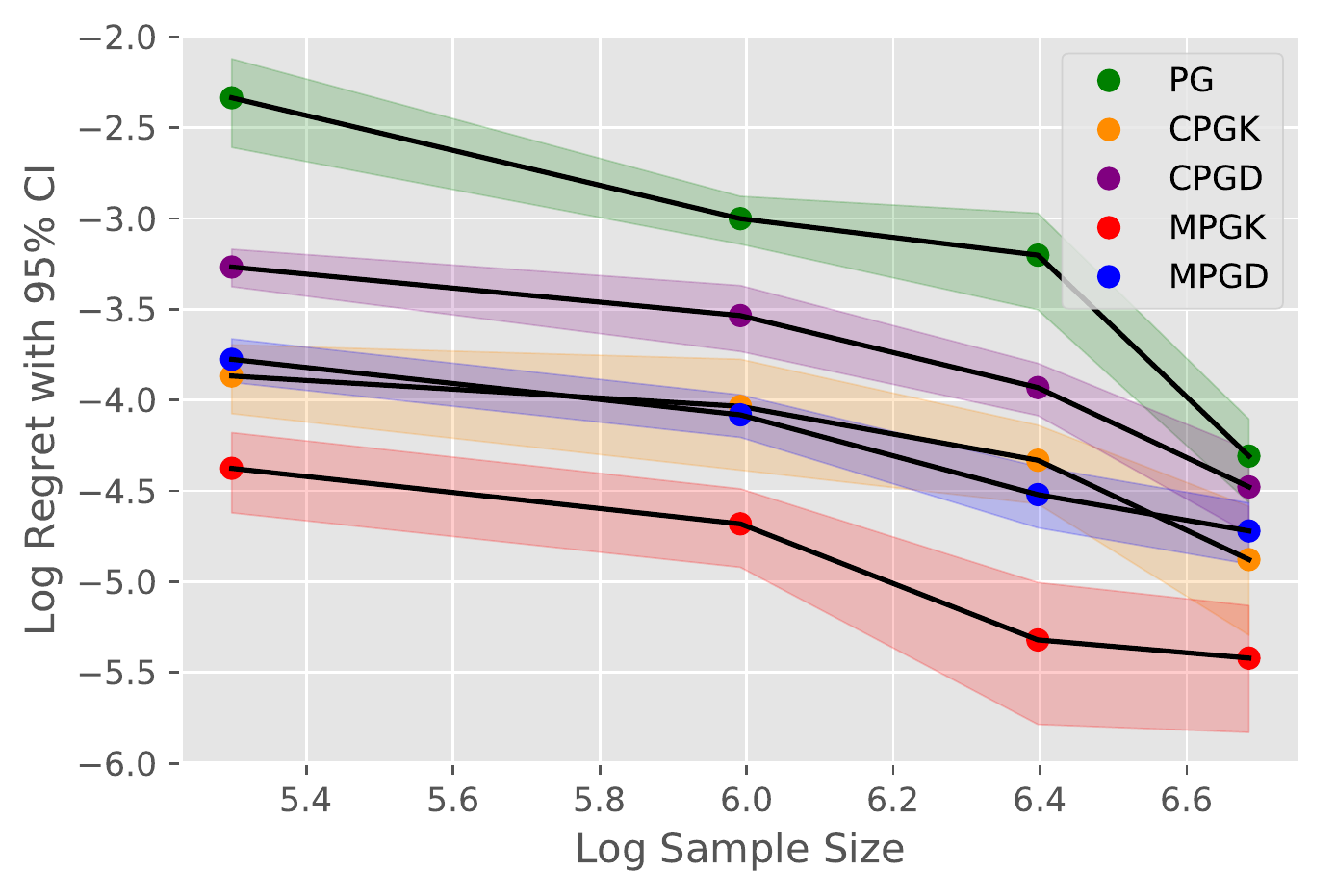}
  \captionof{figure}{Regret after gradient ascent with 95\% CI}
  \label{fig:regret}
\end{minipage}
\end{figure}

First, in \cref{fig:gradient}, we compare the MSE of gradient estimators at $\theta=1.0$ over $100$ replications for each of $n=200,400,600,800$. We find that the performance of MPGK is far superior to all other estimators in terms of MSE, which confirms our theoretical results.
Interestingly, the performance of MPGD is slightly worse than CPGD. The possible reason is it is more difficult to estimate $w$ than $w^{\Kcal}$. The reasonably good performance of CDGD and CDGK can be attributed to the known $\lambda_t^{\Dcal},\lambda_t^{\Kcal}$, which ensures less sensitivity to the $q$-estimation due to the doubly robust error structure.

Second, in \cref{fig:regret}, we apply gradient ascent (see \cref{sec:optimizaiton}) with $\alpha_t=0.05,\,T=50$, and $\hat \theta_1$ randomly chosen from $[0.8,1.2]$. We only run the bootstrap for $\hat \theta_1$ and then keep the same $h$ for the next iterations. We compare the regret of the final policy for the different policy gradient estimators, \ie, $J(\theta^{*})-J(\hat \theta_{50})$, averaging over $100$ replications of the experiment for each of $n=200,400,600,800$. Again, the performance of MPGK is superior to other estimators also in terms of regret.  

\section{Conclusion and Future work}

We developed doubly robust versions of DPG and showed that they can circumvent issues of curse of horizon and of dependence on nuisances such as $q$-estimates.
Theoretically, a next question may be showing the rates we obtain are minimax optimal by appealing to minimax theory for nonparametric density estimation \citep{KorostelevAlexander}.
A more practical next step may be to apply this in larger-scale RL environments. Offline RL in large-scale environments is notoriously difficult \citep{fujimoto19a}. We therefore expect it necessary to combine several heuristics, such as gradient updates to nuisance estimates and adaptive step sizes, to make the algorithm work well in practice.

\bibliographystyle{chicago}
\bibliography{rc}

\appendix 

\newpage 

\section{Notation}
\begin{table}[!h]
    \centering
    \caption{Notation}
    \begin{tabular}{l|l}
    $p_1(s)$ & Initial distributions \\
    $p_{\pi}(\cdot)$ & Induced distribution by a MDP and a policy $\pi$\\
    &$p_\pi(j_{r_{H+1}})= p_1(s_1)\prod_{t=1}^H\pi(a_t\mid s_t)p(r_t\mid s_t,a_t)p(s_{t+1}\mid s_t,a_t)$.\\
     $p_{\tau}(\cdot)$ & $p_\pi(j_{r_{H+1}})= p_1(s_1)\prod_{t=1}^Hp(r_t\mid s_t,\tau_t(s_t))p(s_{t+1}\mid s_t,\tau_t(s_t))$\\
     $p_{\Kcal}(\cdot)$ & $p_\pi(j_{r_{H+1}})=  p_1(s_1)\prod_{t=1}^HK_h(a_t-\tau_t(s_t))p(r_t\mid s_t,a_t)p(s_{t+1}\mid s_t,a_t)$\\     
    $H$ & Horizon \\
    $h$ & Bandwidth \\
    $\traj^{\braangle{i}}$ & i-th data \\
    $\bpol,\epol,\pi^{e,\Kcal}$ & Behavior policy, Evaluation policy, Kernelized policy \\
    $J(\theta),Z(\theta)$ & Value, Gradient \\
    $k(\cdot),K_h(x)$ & Kernel, Normalized kernel $h^{-1}k(h^{-1}x)$ \\
    $\tau,\tau_{\theta}$ & Deterministic policy with a parameter $\theta \in \Theta$ \\ 
    $\E[\cdot]$ & Expectation wrt random variable generated by MDP and a behavior policy \\ 
    $\E_{\epol}[\cdot]$ & Expectation wrt random variable generated by a MDP and a policy $\epol$ \\ 
    $\E_{\epol}[\tau]$ & Expectation wrt random variable generated by a MDP and a policy $\delta(a=\tau_t(s))$ \\ 
    $\ch_{S_t},\ch_{A_t}$ & History $(S_0,A_0,S_1,A_1,\cdots,S_t)$, $(S_0,A_0,S_1,A_1,\cdots,A_t)$   \\
    $h^{u}_{s_t}$ & History $(s_0,u_0,s_1,u_1,\cdots,s_t)$  \\
    $M_2(k)$ & Second moment of kernel, $\int u^2k(u)\rD u$ \\
    $\Omega^{(i)}_{\lambda}(k)$ & Roughness of kernel, $\int k^{2(i)}(u)\rD u$ \\
    $v^{\Kcal},q^{\Kcal}$ & Value, Q-function wrt a kernelized policy and a MDP \\ 
    $v^{\epol},q^{\epol}$ & Value, Q-function wrt a deterministic policy and a MDP  \\
    $q^{(i)}(s,a)$ & $i$-th Differentiation of $q(s,a)$ wrt actions \\ 
    $w^{\Kcal}_t(s_t),w^{\epol}_t(s_t)$  & $p_{\Kcal}(s_t)/p_{\bpol}(s_t),p_{\tau}(s_t)/p_{\bpol}(s_t)$ \\ 
    $\Upsilon$ & $\|\|\otimes \tau_t(S)\|_{\infty}\|_{\oper}\leq \Upsilon$ \\ 
    $R_{\max}$ & $\|R_t\|_{\infty}\leq R_{\max}$ \\
    $C_1$  & $\|1/\bpol(a|s) \|\leq C_1$ \\
    $C_2$ & $\|w^{\epol}_t(s) \|\leq C_2$ \\  
    $\nabla $ & Differentiation wrt $\theta$\\
    $\|\cdot \|_2$    &  $L^2$-norm $\{\E[f^2]\}^{1/2}$  \\
    $\|\cdot \|_{\oper}$    &  Operator norm  \\
    $\|\cdot \|_{i,\infty}$  &  $\max_{j=0,\cdots,i}\|\cdot^{(j)}\|_{\infty} $\\ 
   $d^{w^{\Kcal}}_t(s_t),d^{w^{\epol}}_t(s_t)$ & $\nabla w^{\Kcal}_t(s_t),\nabla w^{\epol}_t(s_t)$ \\ 
   $d^{q^{\Kcal}}_t(s_t,a_t),d^{q^{\epol}}_t(s_t,a_t)$ & $\nabla q^{\epol}_t(s_t,a_t)$ \\ 
   $G^{(i)}_1$ & $\|\int p^{(i))}(r|s,a)\rD r\|_{\infty}\leq G^{(i)}_1$\\
   $G^{(i)}_2$ & $\|\int p^{(i))}(s'|s,a)\rD s'\|_{\infty}\leq G^{(i)}_2$ \\
   $\otimes a$ & $aa^{\top}$ \\
   $f(n)=\bigO(n^a)$ & $f(n)$ is bounded above by $n^a$ asymptotically \\
   $f(n)=\Theta(n^a)$ & $f(n)$ is bounded both above and below by $n^a$ asymptotically \\
  $f(n)=\bigO_p(n^a)$ &  $f(n)/n^a$ is bounded in probability \\
    $f(n)=\smallo_p(n^a)$ & $f(n)/n^a$ converges to $0$ in probability  \\
    $A_n \lessapprox B_n$ & $\exists C, A_n<CB_n$, $C$ is a universal problem independent constant
    \end{tabular}

    \label{tab:my_label}
\end{table}
\newpage

\section{Nuisance estimations}\label{sec:nuisance}

Our algorithm allows any estimators for $q$-functions and marginal ratios to be used. In this section we discuss some standard ways to estimate these nuisance functions. 

\subsection{Estimation of \texorpdfstring{$q$}{a}-functions and their \texorpdfstring{$\theta$}{a}-gradients}

In the tabular case, a model-based approach is the most common way to estimate $q$-functions of $\epol$ from off-policy data. In the non-tabular case, we have to rely on some function approximation. The key equation to derive these methods is the Bellman equation:
\begin{align*}
    q_t(s_t,a_t)&=\E[r_t+q_{t+1}(s_{t+1},\epol)\mid s_t,a_t].
  \end{align*}
There is also an equivalent equation for $d^q$. If $\epol$ is stochastic we may use
\begin{align*}
        d^{q}_j(s_j,a_j) &=  \E[d^{v}_{j+1}(s_{j+1})\mid s_j,a_j],\,d^{v}_j(s_j) =\E_{\epol}[d^{q}_j+g_{j} q_{j}\mid s_{j}], 
\end{align*}
where $q_{t}(s_{t},\pi)=\int q_t(s_t,a_t)\pi(a_t\mid s_t)\mathrm{d}a_t$ and $g_t=\log\epol_t(a_t\mid s_t)$.  When $\epol$ is deterministic, we instead have
\begin{align*}
     d^{q}_j(s_j,a_j) &=  \E[d^{v}_{j+1}(s_{j+1})\mid s_j,a_j],\,d^{v}_j(s_j) =d^{q}_j(s_j,\tau_j(s_j))+q^{(1)}_j(s_j,\tau_j(s_j))\nabla \tau_j(s_j). 
\end{align*}

One of the most common ways to operationalize this is using fitted $q$-iteration (\citealp{antos2008learning,LeHoang2019BPLu}:
\begin{itemize}
    \item Set $\hat q_{H+1}\equiv0$. 
    \item For $t=H,\dots,1$:
    \begin{itemize}
        \item Estimate $\hat q_t$ by regressing $r_t+\hat q_{t+1}(s_{t+1}, \pi^e)$ onto $s_t,a_t$. 
    \end{itemize}
\end{itemize}
Similarly, \citet{EEOPG2020} proposed an analogous estimation method for $d^{q}_j$:
\begin{itemize}
    \item Set $\hat  d^{q}_{H}=0$. 
    \item For $t=H,\dots,1$:
    \begin{itemize}
        \item Estimate $\hat d^{q}_{j}$ by regressing $\hat d^{v}_{j+1}(s_{j+1})$ onto $s_j,a_j$. 
    \end{itemize}
\end{itemize}
The above approaches can be regarded as a dynamic programming approach. When $\epol$ is stochastic, another approach is a Monte Carlo approach based on the equations:
\begin{align*}
   q_j(s_{j},a_{j})  =\E\bracks{\sum_{t=j}^H r_t|s_j,a_j}, \quad d^{q}_j(s_{j},a_{j})=\E\left[\sum_{t=j+1}^{H}r_t\lambda_{j+1:t} \sum_{\ell=j+1}^{t}g_\ell\mid s_{j},a_{j}\right].
\end{align*}
Based on this, we can simply estimate $q$
by regressing $\sum_{t=j}^H r_t$ on $s_j,a_j$ and $d^q$  by regressing  $\sum_{t=j+1}^{H}r_t\lambda_{j+1:t} \sum_{\ell=j+1}^{t}g_\ell$ on $a_{j},s_{j}$. 

\subsection{Estimation of marginal density ratios}

When $\Scal$ is finite, a model based approach \citep{Yin2020} would be a competitive way to estimate marginal density ratios:
\begin{align*}
 \hat  w^{\epol}_t(s_t)=\frac1{\hat p_{\bpol_t}(s_t)}\int \hat p_{t}(s_{t}|s_{t-1},\tau(s_{t-1}))\prod_{k=0}^{t-1}\prns{ \hat p_{k}(s_{k}|s_{k-1},\tau(s_{k-1}))}\mathrm{d}(\ch_{a_{t-1}}), 
\end{align*}
where $\hat p_{{t}},\hat p_{\bpol_t}$ is each an empirical frequency (histogram) estimator. For general state space, we have to rely on some function approximation methods. When the target policy is stochastic, we have the following equations: $j\leq H$, 
\begin{align*}
  w_j(s_j) =\E[\lambda_{j-1}|s_j],\,d^{w}_j =\E\left[\lambda_{0:j-1}\sum_{\ell=0}^{j-1}g_\ell\mid s_j\right]. 
\end{align*}
Thus, for example, $w_j$ is estimated by regressing $\lambda_{j-1}$ onto $s_j$, and $d^{w}_j$ is estimated by regressing $\lambda_{0:j-1}\sum_{\ell=0}^{j-1}g_\ell$ onto $s_j$ \citep{KallusUehara2019,EEOPG2020}. When the evaluation policy is deterministic, it is difficult to estimate directly. 

\section{Off-policy Optimization}\label{sec:optimizaiton}

Our estimated policy gradients can be used in any gradient-based optimization algorithm in order to do off-policy optimization to learn a policy. One example, which we also use in our experiment, is the simple gradient ascent algorithm, given below in \cref{alg:gradient}. Here, $\operatorname{Proj}_{\Theta}$ is the projection onto $\Theta$.

\begin{algorithm}[h!]
 \caption{Off-policy projected gradient ascent}
 \begin{algorithmic}
 \label{alg:gradient}
 \STATE {\bfseries Input:} An initial point $\theta_1\in\Theta$ and step size schedule $\alpha_t$ 
\FOR{$t=1,2,\cdots$}
 \STATE $\tilde \theta_{t+1} =\theta_{t}+\alpha_t \hat Z(\theta_t)$
 \STATE $\theta_{t+1} = \operatorname{Proj}_{\Theta}(\tilde \theta_{t+1})$
 \ENDFOR
\end{algorithmic}
\end{algorithm}

\section{Omitted Proofs}

\subsection{Proof of \texorpdfstring{\cref{thm:ope_final}}{a}}

We prove \cref{thm:ope_final}. Before that, we prove \cref{thm:ope_middle}. 

\begin{theorem}\label{thm:ope_middle}
In the case $\Kcal$, assuming (b), we have 
\begin{align}
\label{eq:replace}
 \E\bracks{\hat J-\E_{n}\left[ \frac{K_h(A-\tau(S))\{R-f_{1}\}}{f_2}+f_3\}\right]} &=\smallo(1/\sqrt{nh}),\\
 \E\bracks{\braces{\hat J-\E_{n}\left[ \frac{K_h(A-\tau(S))\{R-f_{1}\}}{f_2}+f_3\}\right]}^2}&=\smallo(1/(nh)). 
\end{align}
In the case $\Dcal$, the above \eqref{eq:replace} also holds assuming (a).
\end{theorem} 

\subsubsection{Proof of \texorpdfstring{\cref{thm:ope_middle}}{a}, Case \texorpdfstring{$\Kcal$}{a}}

Let us define 
\begin{align*}\ts
    \phi_1(s,a,r;q,\bpol)= \frac{K_h(a-\tau(s))\{r-q(s,a) \}}{\bpol(a\mid s)}+v(s),v(s)=\int K_h(a-\tau(s))q(s,a)\rD (a). 
\end{align*}
As in a standard argument \citep{ChernozhukovVictor2018LRSE,KallusUehara2019}, what we will prove is 
\begin{align}\ts
    \E[\phi_1(S,A,R;\hat q^{*1},\hat \pi^{b*1})-\phi_1(S,A,R;q,\pi^{b})]&= \smallo((nh)^{-1/2}), \label{eq:firstfirst}\\
    \E[\{\phi_1(S,A,R;\hat q^{*1},\hat \pi^{b*1})-\phi_1(S,A,R;q,\pi^{b})\}^2]&= \smallo(h^{-1}). \label{eq:secondsecond}
\end{align}
Then, the same argument holds for $\phi_1(S,A,R;\hat q^{*2},\hat \pi^{b*2})-\phi_1(S,A,R;q,\pi^{b})$. The desired statement is concluded since 
\begin{align*}\ts
&\E[\P_n[ \phi_1(S,A,R;\hat q,\hat \pi^{b}) ]- \P_n[ \phi_1(S,A,R;q,\pi^{b}) ]]\\
&=\E[\E[\P_n[ \phi_1(S,A,R;\hat q,\hat \pi^{b}) ]- \P_n[ \phi_1(S,A,R;q,\pi^{b}) ]\mid \cu_2]] \nonumber,\\
&=\E[\E[\phi_1(S,A,R;\hat q,\hat \pi^{b})-\phi_1(S,A,R;q,\pi^{b})\mid \cu_2]]. 
\end{align*}
In addition, 
\begin{align*}
&\E[\braces{\P_n[ \phi_1(S,A,R;\hat q,\hat \pi^{b}) ]- \P_n[ \phi_1(S,A,R;q,\pi^{b}) ]}^2]\\
&=\E[\E[\braces{\P_n[ \phi_1(S,A,R;\hat q,\hat \pi^{b}) ]- \P_n[ \phi_1(S,A,R;q,\pi^{b}) ]}^2|\cu_2]] \nonumber,\\
&=n^{-1}\E[\E[\{\phi_1(S,A,R;\hat q,\hat \pi^{b})-\phi_1(S,A,R;q,\pi^{b})\}^2\mid \cu_2]]. 
\end{align*}

\paragraph{Proof of \texorpdfstring{\cref{eq:firstfirst}}{a} and \texorpdfstring{\cref{eq:secondsecond}}{a}}

In this subsection, we remove $\{*1\}$ for the ease of the notation. To prove \eqref{eq:firstfirst}, we show $$\E\left[K_h(A-\tau(S))\left\{\frac{1}{\hat \pi^b(A|S)}-\frac{1}{\pi^b(A|S)}\right\}\{q(S,A)-\hat q(S,A)\} \right]=\smallo(nh)^{-1/2}).$$
This is proved by 
{\small 
\begin{align*}\ts
    & \E\bracks{\E\left[K_h(A-\tau(S))\left\{\frac{1}{\hat \pi^b(A|S)}-\frac{1}{\pi^b(A|S)}\right\}\{q(S,A)-\hat q(S,A)\} \mid \cu_2 \right]}\\
    &=\E\bracks{\int \frac{1}{h}k((a-\tau(s))h^{-1}) \left\{\frac{1}{\hat \pi^b(a|s)}-\frac{1}{\pi^b(a|s)}\right\}\{q(s,a)-\hat q(s,a)\}\bpol(a\mid s)p(s)\rD (s,a)}\\
     &=\E\bracks{\int k(u) \left\{\frac{1}{\hat \pi^b(\tau(s)+uh|s)}-\frac{1}{\pi^b(\tau(s)+uh|s)}\right\}\{q(s,\tau(s)+uh)-\hat q(s,\tau(s)+uh)\}\bpol(\tau(s)+uh\mid s)p(s)\rD (u,s)}\\
    &=\E[\int k(u) \left\{\frac{1}{\hat \pi^b(\tau(s)|s)}-\frac{1}{\pi^b(\tau(s)|s)}+uh\braces{-\frac{\hat \pi^{b(1)}(\tau(s)|s)}{\hat \pi^{2b}(\tau(s)|s)}+\frac{\pi^{b(1)}(\tau(s)|s)}{\pi^{2b}(\tau(s)|s)}}+\bigO((uh^2)) \right\}\times\\
    &\{q(s,\tau(s))-\hat q(s,\tau(s))+uh\{q^{(1)}(s,\tau(s))-\hat q^{(1)}(s,\tau(s))\}+\bigO((uh^2))\}\times \\
    &\{\bpol(\tau(s)\mid s)+uh\pi^{b(1)}(\tau(s)\mid s)+\bigO((uh^2))\}p(s)\rD (u,s)]\\
     &=M_2(k)\E\bracks{\int \left\{\frac{1}{\hat \pi^b(\tau(s)\mid s)}-\frac{1}{\pi^b(\tau(s)\mid s)}\right\}\{q(s,\tau(s))-\hat q(s,\tau(s))\}\bpol(\tau(s)\mid s)p(s)\rD (s)}+\bigO(h^2)\times \smallo(1)\\
     &=\smallo((nh)^{-1/2}). 
\end{align*}
}
\normalsize
In the last line, we use the assumptions that $q(a,x),\bpol(a|x),\hat q(a,x),\hat \pi^b(a|x)$ are $C^2$-functions wrt actions, and 
\begin{align*}\ts
    &\E\bracks{\left \|\frac{1}{\hat \pi^b(A\mid S)}- \frac{1}{\pi^b(A\mid S)} \right\|_{\infty}\|q(S,A)-\hat q(S,A) \|_{\infty}}=\smallo(n^{-1/2}h^{-1/2}),\\
    &\E\bracks{\left \|\frac{1}{\hat \pi^b(A)\mid S)}- \frac{1}{\pi^b(A\mid S)} \right\|_{1,\infty}}=\smallo(1),\E\bracks{\|\hat q(S,A)-\hat q(S,A)\|_{1,\infty}}=\smallo(1),\\
    &\bigO(h^2)\times \smallo(1)=\smallo((nh)^{-1/2}),nh^5=\bigO(1). 
\end{align*}

In addition, \cref{eq:secondsecond} is proved since 
\begin{align*}\ts
      &\E[\E[\{\phi_1(S,A,R;\hat q^{*1},\hat \pi^{b*1})-\phi_1(S,A,R;q^{},\pi^{b})\}^2\mid \cu_2]]\\
    &\lessapprox  \E\bracks{\E\left[K_h(A-\tau(S))^2\left\{\frac{1}{\hat \pi^b(A\mid S)}-\frac{1}{\pi^b(A\mid S)}\right\}^2\{q(S,A)-\hat q(S,A)\}^2 \mid \cu_2\right]}  \\
    &+\E\bracks{\E\left[K_h(A-\tau(S))^2\left\{\frac{1}{\hat \pi^b(A \mid S)}-\frac{1}{\pi^b(A \mid S)}\right\}^2 \{R-q(S,A)\}^2\mid \cu_2 \right]} \\
    &+\E\bracks{\E\left[\frac{K_h(A-\tau(S))^2}{\pi^b(A\mid S)^2}\{q(S,A)- \hat q(S,A)\}^2\mid \cu_2  \right]+\E\left[\braces{\hat v(S)-v(S)}^2\mid \cu_2 \right]}  \\
     &\lessapprox h^{-1}\max\left\{\E[\|\hat q(S,\tau(S))-q(S,\tau(S))\|^2_2], \E\bracks{\left \|\frac{1}{\hat \pi^b(\tau(S)|S)}-\frac{1}{\pi^b(\tau(S)|S)}\right\|^2_2} \right\}+\bigO(1)=\smallo(h^{-1}).
\end{align*}

\subsubsection{Proof of \texorpdfstring{\cref{thm:ope_middle}}{a}, Case \texorpdfstring{$\Dcal$}{a}}

Essentially, the same proof is seen in \citet{Kyle2019}. For completeness, we also write the proof here with our notation. Let us define 
\begin{align*}\ts
    \phi_2(s,a,r;q,\bpol)= \frac{K_h(a-\tau(s))\{r-q(s,\tau(s)) \}}{\bpol(a\mid s)}+q(S,\tau(S)). 
\end{align*}
As in a standard argument similar to the case $\Kcal$, what we have to prove is 
\begin{align}\ts
    \E[\phi_2(S,A,R;\hat q^{*1},\hat \pi^{b*1})-\phi_2(S,A,R;q^{},\pi^{b})]=\smallo((nh)^{-1/2}), \label{eq:firstfirst_}\\
    \E[\{\phi_2(S,A,R;\hat q^{*1},\hat \pi^{b*1})-\phi_2(S,A,R;q^{},\pi^{b})\}^2]= \smallo(h^{-1}). \label{eq:secondsecond_}
\end{align}
In this subsection, we remove $\{*1\}$ for the ease of the notation. 

\cref{eq:firstfirst_} is proved since 
\begin{align}\ts
    &\E\bracks{\E[\phi_2(S,A,R;\hat q,\hat \pi^b)-\phi_2(S,A,R;q,\pi^b)\mid \cu_2]} \nonumber \\ 
    &=\E\bracks{\E\left[K_h(A-\tau(S))\left\{\frac{1}{\hat \pi^b(\tau(S)\mid S)}-\frac{1}{\pi^b(\tau(S)\mid S)}\right\}\{q(S,\tau(S))-\hat q(S,\tau(S))\} \mid \cu_2\right]}+  \label{eq:first} \\
    &+\E\bracks{\E\left[K_h(A-\tau(S))\left\{\frac{1}{\hat \pi^b(\tau(S)\mid S)}-\frac{1}{\pi^b(\tau(S)\mid S)}\right\} \{R-q(S,\tau(S))\}\mid \cu_2 \right]} \label{eq:second}\\
    &+\E\bracks{\E\left[\frac{K_h(A-\tau(S))}{\pi^b(\tau(S)\mid S)}\{q(S,\tau(S))- \hat q(S,\tau(S))\}+\hat q(S,\tau(S))-q(S,\tau(S))\mid \cu_2 \right]}    \label{eq:third} \\
     &=\smallo((nh)^{-1/2})+\smallo(1)\times \bigO(h^2)+ \smallo(1)\times \bigO(h^2)=\smallo(nh)^{-1/2}.\nonumber 
\end{align}
Here, we use the facts that \eqref{eq:first} is $\smallo((nh)^{-1/2})$, \eqref{eq:second} is $\smallo(1)\times \bigO(h^2)$, \eqref{eq:third} is $\smallo(1)\times \bigO(h^2)$, which we will prove soon. In the last line, we use $nn^5=\bigO(1)$. From now on, we prove \eqref{eq:second} is $\smallo(1)\times \bigO(h^2)$:
\begin{align*}\ts
    & \E\bracks{[\E\left[K_h(A-\tau(S))\left\{\frac{1}{\hat \pi^b(\tau(S)\mid S)}-\frac{1}{\pi^b(\tau(S)\mid S)}\right\} \{R-q(S,\tau(S))\}\mid \cu_2 \right]} \\
    &=\E\bracks{\E\left[\left\{\frac{1}{\hat \pi^b(\tau(S)\mid S)}-\frac{1}{\pi^b(\tau(S)\mid S)}\right\}\{\E[K_h(A-\tau(S))q(S,A)\mid S]-K_h(A-\tau(S))q(S,\tau(S))\}\mid \cu_2  \right]} \\ 
 &=\E\bracks{\E\left[\left\{\frac{1}{\hat \pi^b(\tau(S)\mid S)}-\frac{1}{\pi^b(\tau(S)\mid S)}\right\}\left\{\bigO(h^2)\right\} \mid \cu_2  \right]} \\
 &= \smallo(1)\times \bigO(h^2).
\end{align*}
More  specifically,
\begin{align*}
    &\E[K_h(A-\tau(S))\{q(S,A)-q(S,\tau(S))\}\mid S]=\int \frac{1}{h}k\braces{\frac{a-\tau(s)}{h}}\bpol(a|s)\{q(s,a)-q(s,\tau(s))\}\mathrm{d}a\\
    &=\int k(u)\bpol(\tau(s)+uh|s)\{q(s,\tau(s)+uh)-q(s,\tau(s))\}\mathrm{d}u\\
        &=\int k(u)\{\bpol(\tau(s)|s)+\bigO(uh)\}\{uh q^{(1)}(s,\tau(s))+\bigO(h^2) \}\mathrm{d}u=\bigO(h^2). 
\end{align*}
noting $\int uk(u)\mathrm{d}u=0$. Next, we prove \eqref{eq:third} is $\smallo(1)\times \bigO(h^2)$: 
\begin{align*}\ts
&\E\bracks{\E\left[\frac{K_h(A-\tau(S))}{\pi^b(\tau(S)\mid S)}\{q(S,\tau(S))- \hat q(S,\tau(S))\}+\hat q(S,\tau(S))-q(S,\tau(S))\mid \cu_2 \right]}\\
&=\E\bracks{\E\left[\left\{\frac{K_h(A-\tau(S))}{\pi^b(\tau(S)\mid S)}-1\right\}\{q(S,\tau(S))- \hat q(S,\tau(S))\}\mid \cu_2 \right]}\\ 
&=\E\bracks{\E\left[\{\bigO(h^2)\}\{q(S,\tau(S))- \hat q(S,\tau(S))\}\mid \cu_2 \right]}=\smallo(1)\times \bigO(h^2). 
\end{align*}

\cref{eq:secondsecond_} is similarly proved as in the case $\Kcal$.

\subsubsection{Proof of \texorpdfstring{\cref{thm:ope_final}}{a}}
We prove the statement for the case $\Kcal$. The statement for the case $\Dcal$ is similarly proved as in \citep{Kyle2019}. 
~
\paragraph{Bias term }
The bias term is calculated as follows:
\begin{align*}\ts
&\E[\P_n[\phi_1(S,A,R;q,\pi^b)]]-J \\
&=\E\left[\int K_h(a-\tau(S))q(S,a)\rD a\right]-J \\ 
&= \E\left[\int k(u)\{q(S,\tau(S)+uh)-q(S,\tau(S))\}\rD u\right] \\ 
&= 0.5h^2\E\left[\int k(u)u^2q^{(2)}(S,\tau(S))\rD u\right]+o(h^2)= 0.5h^2M_2(k)\E\left[q^{(2)}(S,\tau(S))\right]+o(h^2). 
\end{align*}
Here, we use a smoothness assumption. More formally, from the third line to the fourth line, based on the function $a \to q(s,a)$ is a $C^2$-function on the compact space, we use
\begin{align*}\ts
    q(s,\tau(s)+uh)-q(s,\tau(s))=uhq^{(1)}(s,\tau(s))+0.5(uh)^2q^{(2)}(s,\tau(s))+o((uh)^2). 
\end{align*}
Refer to \citet[Exericise 1.5]{LiQi2007Ne:t}. Then, the all of the bias is 
\begin{align*}\ts
    \E[\P_n[\phi_1(S,A,R;q,\pi^b)]]-J &= \E[\P_n[\phi_1(S,A,R;q,\pi^b)]]-J+\smallo(n^{-1/2}h^{-1/2})\\ 
    &=0.5h^2M_2(k)\E\left[q^{(2)}(S,\tau(S))\right]+\smallo(n^{-1/2}h^{-1/2}). 
\end{align*}
Finally, noting $\hat J^{\Kcal }-J=\E[\P_n[\phi_1(S,A,R;q,\pi^b)]]-J+\smallo((nh)^{-1/2})$, the statement is concluded. 

\paragraph{Variance term }

The variance term is calculated as follows. First, we have 
\begin{align*}\ts
    &\mathrm{var}[\P_n[\phi_1(S,A,R;q,\pi^b)]] \\
    &= \frac{1}{n}\left(\E[\phi_1(S,A,R;q,\pi^b)^2]- \{\E[\phi_1(S,A,R;q,\pi^b)]\}^2\right)\\
     &= \frac{1}{nh^2}[\int  \left\{\frac{k((a-\tau(s))h^{-1})}{\pi^b(a \mid s)}\right\}^2\{r-q(s,a)\}^2p(r\mid a,s)\bpol(a\mid s)p(s)\rD (a,s,r) \\
     &+\int \left\{\int  k((a-\tau(s))h^{-1})q(s,a)\rD a\right\}^2 p(s)\rD (s) +O(h^2)]\\
   &= \frac{1}{nh^2}\left[\int  h\left\{\frac{k^2(u)}{\pi^b(\tau(s)+uh \mid s)}\right\}\{r-q(s,\tau(s)+uh)\}^2p(r\mid s,\tau(s)+uh)p(s)\rD (u,s,r) +O(h^2)\right] \\
   &= \frac{1}{nh^2}\left[\int  h\left\{\frac{k^2(u)}{\pi^b(\tau(s)\mid s)}\right\}\{r-q(s,\tau(s))\}^2p(r\mid s,\tau(s))p(s)\rD (u,s,r) +O(h^2)\right]\\
   &=  \frac{1}{nh}\{\Omega_2(k) V+o(h)\}, \,
  V= \int \left\{\frac{\{r-q(s,\tau(s))\}^2}{\pi^b(\tau(s)\mid s)}\right\}p(r\mid s,\tau(s))p(s)\rD (s,r). 
\end{align*} 
Here, we use smoothness assumptions, and 
$\left\{\int  k((a-\tau(s))h^{-1})q(s,a)\rD a\right\}^2=O(h^2)$, which is proved by a standard algebra. 
Then, 
\begin{align*}\ts
    &\mathrm{var}[\P_n[\phi_1(S,A,R;\hat q,\hat \pi^b)]]\\
    &=\mathrm{var}[\P_n[\phi_1(S,A,R;q,\pi^b)]]+\mathrm{var}[\P_n[\phi_1(S,A,R;q,\pi^b)-\phi_1(S,A,R;\hat q,\hat \pi^b)]]\\
    &+2\{\mathrm{var}[\P_n[\phi_1(S,A,R;q,\pi^b)]]\mathrm{var}[\P_n[\phi_1(S,A,R;q,\pi^b)-\phi_1(S,A,R;\hat q,\hat \pi^b)]] \}^{1/2} \\
    &=  \frac{\Omega_2(k)}{nh}\{V+o(h)\}+2\{\frac{1}{nh}\{V+o(h)\}\}^{1/2}\smallo(n^{-1/2}h^{-1/2})+\smallo(n^{-1}h^{-1})\\ 
    &=  \frac{\Omega_2(k)}{nh}\{V+\smallo(1)\}. 
\end{align*}
\begin{remark}
\citet[Theorem 1]{Kyle2019} showed that the constant in the bias term is 
\begin{align*}
   \E[0.5q^{(2)}(s,\tau(s))+q^{(1)}(s,\tau(s))\pi^{b(1)}(a|s)/\pi^{b}(a|s)]. 
\end{align*}
\end{remark}

\subsection{Proof of \texorpdfstring{\cref{cor:optimal}}{a}}

Obvious from \cref{thm:ope_final}. 

\subsection{Proof of \texorpdfstring{\cref{thm:theory_pg}}{a}}

~
\paragraph{Replacing parts}

Let us define 
\begin{align*}\ts
    \psi^{\Kcal}_1(s,a,r;q,\bpol)= \braces{\frac{\nabla K_h(a-\tau(s))\{r-q(s,a) \}}{\bpol(a\mid s)}+\int \nabla K_h(a-\tau(s))q(s,a)\rD (a)}. 
\end{align*}
Here, we prove that nuisance estimators can be replaced with true functions in the sense that 
\begin{align*}\ts
&\E[A_n]= \smallo(n^{-1/2}h^{-3/2}),\,\E[A^2_n]= \smallo(n^{-1}h^{-3}), 
\end{align*}
where 
\begin{align*}
A_n=\P_{\cu_1}[\psi^{\Kcal}_1(S,A,R;\hat q^{*1},\hat \pi^{b*1})|\cu_2]+\P_{\cu_2}[\psi^{\Kcal}_1(S,A,R;\hat q^{*2},\hat \pi^{b*2})|\cu_1]-\P_n[\psi^{\Kcal}_1(S,A,R; q,\bpol)] . 
\end{align*}

Then, what we have to prove is 
\begin{align}\ts
    \E[\E[\psi^{\Kcal}_1(S,A,R;\hat q^{*1},\hat \pi^{b*1})-\psi^{\Kcal}_1(S,A,R;q,\pi^b)\mid \cu_2] ]&= \smallo(n^{-1/2}h^{-3/2}) \label{eq:dcal1},\\ 
    \E[\E[\{\psi^{\Kcal}_1(S,A,R;\hat q^{*1},\hat \pi^{b*1})-\psi^{\Kcal}_1(S,A,R;q,\pi^b)\}^2\mid \cu_2]]&= \smallo(h^{-3})\label{eq:dcal2}. 
\end{align}
In this subsection, we remove $\{*1\}$ for the ease of the notation. We write $1/\bpol(A|S)$ as $\eta(A|S)$. 
\cref{eq:dcal1} is proved as follows:
{\small 
\begin{align}\ts
   & \E[\E[\psi^{\Kcal}_1(S,A,R;\hat q,\hat \pi^b)-\psi^{\Kcal}_1(S,A,R;q,\pi^b)\mid \cu_2]]\\
  &=\E[\E\left[\nabla K_h(A-\tau(S))\left\{\hat \eta(A|S)-\eta(A|S)\right\}\{q(S,A)-\hat q(S,A)\} \mid \cu_2 \right] ] \nonumber \\
&=\E\bracks{\E\left[-h^{-2}k^{(1)}\left(\frac{A-\tau(S)}{h}\right)\nabla \tau(S)\left\{\hat \eta(A|S)-\eta(A|S)\right\}\{q(S,A)-\hat q(S,A)\} \mid \cu_2 \right]}   \nonumber\\
&=\E\bracks{\E\left[-h^{-2}k^{(1)}\left(\frac{A-\tau(S)}{h}\right)\nabla \tau(S)\left\{\hat \eta(A|S)-\eta(A|S)\right\}\{q(S,A)-\hat q(S,A)\} \mid \cu_2 \right] }  \nonumber\\  
&=\E[\E[\int -h^{-1}k^{(1)}\left(u\right)\nabla \tau(S)\left\{\hat \eta(\tau(S)+uh|S)-\eta(\tau(S)+uh|S)\right\}\times \nonumber \\
&\{q(S,\tau(S)+uh)-\hat q(S,\tau(S)+uh)\}\bpol(\tau(s)+uh|S)\rd u \mid \cu_2]]  \label{eq:pre1}\\
&=\E[\E[\int k\left(u\right)\nabla \tau(S)\left\{\hat \eta^{(1)}(\tau(S)+uh|S)-\eta^{(1)}(\tau(S)+uh|S)\right\}\times \nonumber \\
&\{q(S,\tau(S)+uh)-\hat q(S,\tau(S)+uh)\}\pi^{b}(\tau(S)+uh|S)\rd u \mid \cu_2]] \label{eq:pre2}\\
&+\E[\E[\int k\left(u\right)\nabla \tau(S)\left\{\hat \eta(\tau(S)+uh|S)-\eta(\tau(S)+uh|S)\right\}\times \nonumber \\ 
&\{q^{(1)}(S,\tau(S)+uh)-\hat q^{(1)}(S,\tau(S)+uh)\}\pi^{b}(\tau(S)+uh|S)\rd u \mid \cu_2]]  \nonumber \\
&+\E[\E[\int k\left(u\right)\nabla \tau(S)\left\{\hat \eta(\tau(S)+uh|S)-\eta(\tau(S)+uh|S)\right\}\times \nonumber \\
&\{q(S,\tau(S)+uh)-\hat q(S,\tau(S)+uh)\}\pi^{b(1)}(\tau(S)+uh|S)\rd u \mid \cu_2] ] \nonumber 
\end{align}
Then, this is equal to  
\begin{align}\ts
 &M_2(k)\E[\E[\nabla \tau(S)\left\{\hat \eta^{(1)}(\tau(S)|S)-\eta^{(1)}(\tau(S)|S)\right\}\{q(S,\tau(S))-\hat q(S,\tau(S))\}\bpol(\tau(S)|S)\mid \cu_2] ]\label{eq:pre3} \\
&+M_2(k)\E[\E[\nabla \tau(S)\left\{\hat \eta(\tau(S)|S)-\eta(\tau(S)|S)\right\}\{q^{(1)}(S,\tau(S))-\hat q^{(1)}(S,\tau(S))\}\bpol(\tau(S)|S)\mid \cu_2] ] \nonumber \\
&+M_2(k)\E[\E[\nabla \tau(S)\left\{\hat \eta(\tau(S)|S)-\eta(\tau(S)|S)\right\}\{q^{}(S,\tau(S))-\hat q^{}(S,\tau(S))\}\bpol(\tau(S)|S)\mid \cu_2] ]  \nonumber \\
&+\smallo(h^2) \nonumber \\
&\lessapprox \E[\|\|\nabla \tau(S)\|_{\oper}\|_{\infty}\|\hat q(S,A)-q(S,A)\|_{1,\infty} \|\hat \pi^b(S,A)-\pi^b(S,A)\|_{1,\infty}] + \smallo(n^{-1/2}h^{-3/2})\label{eq:pre4}\\
&= \smallo(n^{-1/2}h^{-3/2}). \nonumber
\end{align}
}
Here, from \eqref{eq:pre1} to \eqref{eq:pre2}, we have used a partial integration. From \eqref{eq:pre2} to \eqref{eq:pre3}, we have used $a\to \eta(s,a)$ and $a\to q(s,a)$ are $C^3$-functions, and 
{\small 
\begin{align*}\ts
    &\E\bracks{\left\{\hat \eta^{(1)}(\tau(S)+uh|S)-\eta^{(1)}(\tau(S)+uh|S)\right\}\{q(S,\tau(S)+uh)-\hat q(S,\tau(S)+uh)\}\mid \cu_2}\\
    &=\E[\left\{\hat \eta^{(1)}(\tau(S)|S)-\eta^{(1)}(\tau(S)|S)+uh\{\hat \eta^{(2)}(\tau(S)|S)-\eta^{(2)}(\tau(S)|S) \}+\bigO((uh)^2)\right\}\times \\
    &\{q(S,\tau(S))-\hat q(S,\tau(S))+uh\{q^{(1)}(S,\tau(S))-\hat q^{(1)}(S,\tau(S))\}+\bigO((uh)^2)\}\mid \cu_2]\\
    &\lessapprox \|\hat \eta(A|S)-\eta(A|S)\|_{1,\infty}\times  \|\hat q(A|S)-q(A|S)\|_{1,\infty}\\ 
    &+\{\|\hat \eta(A|S)-\eta(A|S)\|_{2,\infty} \|\hat q(A|S)-q(A|S)\|_{2,\infty}+\|\hat \eta(A|S)-\eta(A|S)\|_{1,\infty}+\|\hat \eta(A|S)-\eta(A|S)\|_{1,\infty}\}\times \bigO(h^2). 
\end{align*}
}
From \eqref{eq:pre3} to \eqref{eq:pre4},  we use an assumption $nh^7=\bigO(1)$. 

\cref{eq:dcal2} is proved as follows:

\begin{align*}\ts
   &\E\bracks{\E[\{\psi^{\Kcal}_1(S,A,R;\hat q,\hat \eta)-\psi^{\Kcal}_1(S,A,R;q,\eta)\}^2 \mid \cu_2]} \\
    &\lessapprox  \E\bracks{\E\left[h^{-2}K^{(1)}_h(A-\tau(S))^2\left\{\frac{1}{\hat \pi^b(A\mid S)}-\frac{1}{\pi^b(A\mid S)}\right\}^2\{q(S,A)-\hat q(S,A)\}^2 \{\nabla \tau(S)\}^2 \mid \cu_2\right]}  \\
    &+  \E\bracks{\E\left[h^{-2}K^{(1)}_h(A-\tau(S))^2\left\{\frac{1}{\hat \pi^b(A \mid S)}-\frac{1}{\pi^b(A \mid S)}\right\}^2 \{R-q(S,A)\}^2\{\nabla \tau(S)\}^2 \mid \cu_2 \right]} \\
    &+  \E\bracks{\E\left[h^{-2}\frac{K^{(1)}_h(A-\tau(S))^2}{\pi^b(A\mid S)^2}\{q(S,A)- \hat q(S,A)\}^2\{\nabla \tau(S)\}^2 \mid \cu_2  \right]}\\
    &+ \E\bracks{\E\left[h^{-2}\braces{\int K^{(1)}_h(a-\tau(S))\hat q(S,a)\mathrm{d}a  -\int K^{(1)}_h(a-\tau(S))q(S,a)\mathrm{d}a}^2\{\nabla \tau(S)\}^2 \mid \cu_2 \right]}   \\
   &\lessapprox  h^{-3}\times \E[\max\{\|\hat \eta(S,\tau(S))-\eta(S,\tau(S))\|^2_{2},\|\hat \eta^{(1)}(S,\tau(S))-\eta^{(1)}(S,\tau(S))\|^2_2, \\
   &\|\hat q(S,\tau(S))-q(S,\tau(S))\|^2_2,\|\hat q^{(1)}(S,\tau(S))-q^{(1)}(S,\tau(S))\|^2_2\}]+\smallo(h^{-2})\\
   &=\smallo(h^{-3}). 
\end{align*}

\paragraph{Calculation of the bias and variance term}

The bias term is calculated as 
\begin{align*}\ts
    &\E[\P_n[\psi^{\Kcal}_1(S,A,R;q,\pi^b)]]-Z({\theta})\\
    &=\E\left[-\frac{\nabla \tau_{\theta}(S)}{h^2}\int k^{(1)}\left(\frac{a-\tau_{\theta}(S)}{h}\right)q(S,a)\rD a\right]-Z({\theta})\\
    &=\E\left[-\frac{\nabla \tau_{\theta}(S)}{h}\int k^{(1)}\left(u\right)q(S,\tau_{\theta}(S)+uh)\rD u\right]-Z({\theta})\\
     &=\E\left[\nabla \tau_{\theta}(S)\int k\left(u\right)q^{(1)}(S,\tau_{\theta}(S)+uh)\rD u\right]-Z({\theta})\\
 &=\E\left[\nabla \tau_{\theta}(S)\int k\left(u\right)\{q^{(1)}(S,\tau_{\theta}(S)+uh)-q^{(1)}(S,\tau_{\theta}(S))\} \rD u\right]\\
  &=0.5h^2\E\left[\nabla \tau_{\theta}(S)q^{(3)}(S,\tau_{\theta}(S))\right] \int u^2 k(u)\rD u+\smallo(h^2). 
\end{align*}
In the last line, we have used that the function $a \to q(s,a)$ is a $C^3$-function. In addition, we also use a fact $nh^7=\bigO(1)$ to say 
\begin{align*}\ts
    \hat Z-Z({\theta})&=\E[\P_n[\psi^{\Kcal}_1(S,A,R;q,\pi^b)]]-Z({\theta})+\smallo(h^2)\\
      &=0.5h^2\E\left[\nabla \tau_{\theta}(S)q^{(3)}(S,\tau_{\theta}(S))\right] \int u^2 k(u)\rD u+\smallo(n^{-1/2}h^{-3/2}).
\end{align*}
\begin{remark}
Heuristically, this $0.5\E\left[\nabla \tau_{\theta}(S)q^{(3)}(S,\tau_{\theta}(S))\right]$ 
is calculated by differentiating the bias term of the OPE estimator: 
$\E\left[q^{(2)}(S,\tau_{\theta}(S))\right]$. 
\end{remark}
The variance term is calculated as 
\begin{align*}\ts
    &\mathrm{var}[\P_n[\psi^{\Kcal}_1(S,A,R;q,\pi^b)]]=\frac{1}{n}\E\left[\psi^{\Kcal}_1(S,A,R;q,\pi^b)^2 \right]+\op(\frac{1}{nh^3})\\
    &= \frac{1}{n}\E\left[\otimes \nabla \tau_{\theta}(S) \frac{ \{R-q(S,A)\}^2 }{\{\pi^b(A\mid S)\}^2} K^{(1)}_h(A-\tau_{\theta}(S))^2\right]+\smallo(\frac{1}{nh^3}) \\
    &=\frac{1}{n}\E[\otimes \nabla \tau_{\theta}(S)\int \frac{\{r-q(S,a)\}^2}{\pi^b(a\mid S)} K^{(1)}_h(a-\tau_{\theta}(S))^2 p(r\mid S,a)\rD(a,r) ] \\
    &=\frac{1}{nh^4}\E[\otimes \nabla \tau_{\theta}(S)\int \frac{\{r-q(S,a)\}^2}{\pi^b(a\mid S)} k^{(1)}\left(\frac{a-\tau_{\theta}(S)}{h}\right)^2 p(r\mid S,a)\rD(a,r) ]+\smallo(\frac{1}{nh^3})\\
    &=\frac{1}{nh^3}\E[\otimes \nabla \tau_{\theta}(S)\int \frac{\{r-q(S,\tau_{\theta}(S)+hu)\}^2}{\pi^b(\tau_{\theta}(S)+hu\mid s)} k^{(1)}\left(u\right)^2 p(r \mid S,\tau_{\theta}(S)+hu )\rD(u,r) ]+\smallo(\frac{1}{nh^3})\\
&=\frac{1}{nh^3}\left\{\E[\otimes \nabla \tau_{\theta}(S)\int \frac{\{r-q(S,\tau_{\theta}(S))\}^2}{\pi^b(\tau_{\theta}(S)\mid S)} k^{(1)}\left(u\right)^2 p(r\mid S,\tau_{\theta}(S))\rD(u,r) ]+\smallo(1)\right\}\\
&=\frac{1}{nh^3}\int k^{(1)}\left(u\right)^2  \rD (u) \left\{\E\left[\otimes \nabla \tau_{\theta}(S) \frac{\mathrm{var}[R\mid S,\tau_{\theta}(S)]}{\pi^b(\tau_{\theta}(S)\mid S)}\right] +\smallo(1)\right\}. 
\end{align*}
Thus, 
\begin{align*}\ts
    \mathrm{var}[\P_n[\psi^{\Kcal}_1(S,A,R;\hat q,\hat \pi^b)]] &=\mathrm{var}[\P_n[\psi^{\Kcal}_1(S,A,R;q,\pi^b)]]+\smallo(n^{-1}h^{-3})\\
    &= \frac{1}{nh^3}\Omega^{(1)}_2(k)\left\{\E\left[\otimes \nabla \tau_{\theta}(S) \frac{\mathrm{var}[R\mid S,\tau_{\theta}(S)]}{\pi^b(\tau_{\theta}(S)\mid S)}\right] +\smallo(1)\right\}. 
\end{align*}

\subsection{Proof of \texorpdfstring{\cref{thm:OPE_RL}}{a}}

\paragraph{Replacing estimators with true functions}

We define 
$$\phi^{\Kcal}(\bpol, q^{\Kcal})=v^{\Kcal}_0+\sum_{t=1}^H \lambda^{\Kcal}_t\{r_t-q^{\Kcal}_t+v^{\Kcal}_{t+1}\}.$$
Here, we prove that nuisance estimators can be replace with true functions:
\begin{align*}\ts
    \P_{\cu_1}[\phi^{\Kcal}(\hat \pi^{b\Kcal*1}, \hat q^{\Kcal*1})|\cu_2] +  \P_{\cu_2}[\phi^{\Kcal}(\hat \pi^{b\Kcal*2}, \hat q^{\Kcal*2})|\cu_1] &=    \E[\phi^{\Kcal}(\bpol,q^{\Kcal})]+\op((nh^H)^{-1/2}).
\end{align*}
Then, what we have to prove is 
\begin{align*}\ts
   \E[ \phi^{\Kcal}( \hat \pi^b, \hat q^{\Kcal})-\phi^{\Kcal}(\bpol, q^{\Kcal}) |\cu_2 ]  &=\op((nh^H)^{-1/2}), \\
    \E[ \{\phi^{\Kcal}( \hat \pi^b, \hat q^{\Kcal})-\phi^{\Kcal}(\bpol, q^{\Kcal}) \}^2 |\cu_2] &=\op(h^{-H}) . 
\end{align*}
The rest of the part is proved as \cref{thm:rl_ope2}. Therefore, we omit the proof here. 

Next, we analyze the bias and variance. 
~
\paragraph{Bias part}

First, we have 
\begin{align*}\ts
    \E[\P_n[\phi^{\Kcal}(\bpol,q^{\Kcal})]]=\E[\sum_t \lambda^{\Kcal}_t r_t]-J(\theta)=\sum_t \E[\{\lambda^{\Kcal}_t-\lambda_t\} r_t].
\end{align*}   
Here, we use a doubly robust property of $\phi^{\Kcal}$. Then, by defining $c=q_t(s_t,a_t)$, the above is equal to 
{\small 
\begin{align*}\ts   
    &\sum_t \left\{\int \E[R_t \mid S_t=s_t,A_t=a_t]\prod_{i=1}^{t}\frac{K_h(a_i-\tau({s_i}))}{\bpol(a_i\mid {s_i})} \{   \prod_{i=1}^{t}\bpol(a_i\mid {s_i})p(s_i \mid s_{i-1},a_{i-1}) \}\rD (h_{a_t})-\E[\lambda_t r_t]\right\}\\
     &=\sum_t \left\{\int c(s_t,\tau_t(s_t)+hu_t)\prod_{i=1}^{t}k(u_i) \prod_{i=1}^{t}p(s_i \mid s_{i-1},\tau(s_{i-1})+hu_i) \rD (h^{u}_{a_t})-\E[\lambda_t r_t]\right\} \\
   &=\sum_t \{\int \{c(s_t,\tau_t(s_t)+hu_t)-c(s_t,\tau_t(s_t))\}\prod_{i=1}^{t}k(u_i) \prod_{i=1}^{t}p(s_i \mid s_{i-1},\tau(s_{i-1})+hu_i) \rD (h^{u}_{a_t})-\\
   &+\int c(s_t,\tau_t(s_t))\prod_{i=1}^{t}k(u_i) \prod_{i=1}^{t}\{p(s_i \mid s_{i-1},\tau(s_{i-1})+hu_i)-p(s_i \mid s_{i-1},\tau(s_{i-1}))\} \rD (h^{u}_{a_t}) \}
\end{align*}
}
where $c(s_t,a_t)=\E[Y_t \mid s_t,a_t],\,h^{u}_{a_t}=\{s_1,u_1,s_2,\cdots \},\,h^{u\tau}_{a_t}=\{s_1,\tau(s_1)+hu_1,s_2, \cdots \},\\ h^{\tau}_{a_t}=\{s_1,\tau(s_1),s_2,\cdots \}$.
Then, we have 
\begin{align*}\ts
 &\sum_t \{\int \{h^2u^2_t c^{(2)}_t(s_t,\tau_t(s_t))\}\left\{\prod_{i=1}^{t}k(u_i)  p(s_i \mid s_{i-1},
 \tau(s_{i-1}))\right \}\rD (h^{u}_{a_t}) -\\
   &\int c(s_t,a_t)\left\{\prod_{i=1}^{t}k(u_i)\right\}\sum_{l=1}^{t-1} \{h^2u^2_l p^{(2)}(s_{l+1} \mid s_l, \tau(s_l)) \prod_{j\neq l} p(s_j\mid s_{j-1},\tau(s_{j-1}) )\}\rD (h^{u}_{a_t}) \}+o(h^2).
\end{align*}
Here, $h^{-u}_{a_t}=\{s_1,s_2,\cdots \}$. Finally, it is equal to 
\begin{align*}\ts
    0.5h^2M_2(k)\sum_{t=1}^{H}\left\{\E_{\tau}[\frac{r_tp^{(2)}(r_t|s_t,\tau_t(s_t))}{p(r_t|s_t,\tau_t(s_t))} ] +\sum_{l=1}^{t-1} \E_{\tau}\left[\frac{r_tp^{(2)}(s_{l+1}\mid s_l,\tau(s_l))}{p(s_{l+1}\mid s_l,\tau(s_l))}\right ] \right\}+\smallo(h^2).
\end{align*}
Then, we have 
{\small 
\begin{align*}\ts
&\E[\phi^{\Kcal}(\hat \pi^b,\hat q^{\Kcal})]=\E[\phi^{\Kcal}(\bpol,q^{\Kcal})]+\op((nh^H)^{-1/2})\\
    &=0.5h^2M_2(k)\sum_{t=1}^{H}\left\{\E_{\tau}\bracks{\frac{r_tp^{(2)}(r_t|s_t,\tau_t(s_t))}{p(r_t|s_t,\tau_t(s_t))} } +\sum_{l=1}^{t-1} \E_{\tau}\left[\frac{r_tp^{(2)}(s_{l+1}\mid s_l,\tau(s_l))}{p(s_{l+1}\mid s_l,\tau(s_l))}\right ] \right\}+o(h^2)+\op((nh^H)^{-1/2})\\
        &=0.5h^2M_2(k)\sum_{t=1}^{H}\left\{\E_{\tau}\bracks{\frac{r_tp^{(2)}(r_t|s_t,\tau_t(s_t))}{p(r_t|s_t,\tau_t(s_t))} } +\sum_{l=1}^{t-1} \E_{\tau}\left[\frac{r_tp^{(2)}(s_{l+1}\mid s_l,\tau(s_l))}{p(s_{l+1}\mid s_l,\tau(s_l))}\right ] \right\}+\op((nh^H)^{-1/2}). 
\end{align*}
}
~
\paragraph{Variance part}

The variance is 
{\small 
\begin{align}\ts
    &\mathrm{var}[\P_n[\phi^{\Kcal}(\bpol,q^{\Kcal})]] \nonumber =\frac{1}{n}\mathrm{var}[\phi^{\Kcal}(\bpol,q^{\Kcal})]] \nonumber \\
     &=\frac{1}{n}\sum_{t=1}^{H} \E\left[\left\{\prod^t_{i=1}\frac{h^{-1}k(h^{-1}(A_i-S_i))}{\bpol(A_i \mid S_i)}\right\}^2\mathrm{\var}[R_t+v^{\Kcal}(S_{t+1})\mid A_t,S_t] \right]. \label{eq:variance} 
\end{align}
}
Here, $\forall f$, we have  
\begin{align*}\ts
    &\E\left[ \left\{\prod^t_{i=1}\frac{h^{-1}k(h^{-1}(A_i-\tau({S_i})))}{\bpol(A_i \mid {S_i})}\right\}^2 f(\ch_{A_t})\right] \\
    &= \frac{1}{h^{2t}}\int \prod^t_{i=1}\frac{k^2(h^{-1}(a_i-\tau({s_i})))}{\bpol(a_i\mid s_i) }f(h_{a_t})\{\prod^{t}_{i=1} p(s_i\mid s_{i-1},a_{i-1}) \} \rD (h_{a_t}) \\
    &= \frac{1}{h^{t}}\int \prod^t_{i=1}\frac{k^2(u_i)}{\bpol(hu_i+\tau_i(s_i)\mid s_i) }f(h^{u\tau}_{a_t})\{\prod^{t}_{i=1} p(s_i\mid s_{i-1},hu_{i-1}+\tau(s_{i-1}) \} \rD (h^{u}_{a_t}) \\
    &= \frac{1}{h^{t}}\left\{\int \prod^t_{i=1}\frac{k^2(u_i)}{\bpol(\tau_i(s_i)\mid {s_i}) }f(h^{u\tau}_{a_t})\{\prod^{t}_{i=1} p(s_i\mid s_{i-1},\tau(s_{i-1})) \}\rD (h^{u}_{a_t})+o(1)\right\}\\
    &= \frac{1}{h^{t}}\left\{\prod_{i=1}^{t} \int k^2(u_i)\rD (u_i)\right\}\left\{\int \prod^t_{i=1}\left\{\frac{1}{\bpol(\tau({s_i})\mid {s_i}) }\right\}f(h^{\tau}_{a_t})\{\prod^{t}_{i=1} p(s_i\mid  s_{i-1},\tau(s_{i-1})) \}\rD (h^{\tau}_{a_t})+o(1)\right\}.
\end{align*}
Therefore, \eqref{eq:variance} is 
\begin{align*}\ts
     \frac{1}{nh^{H}}\left\{\left\{\prod_{i=1}^{H} \int k^2(u_i)\rD (u_i)\right\}\E_{\tau}\left[\prod^H_{i=1}\left\{\frac{1}{\bpol(\tau_i(s_i)\mid s_i)} \right\}\var[r_H|s_H,\tau(s_H)] \right]+\op(1)\right\}, 
\end{align*}
Finally, 
\begin{align*}\ts
    \mathrm{var}[\P_n[\phi^{\Kcal}(\hat \pi^b,\hat q^{\Kcal})]]&= \mathrm{var}[\P_n[\phi^{\Kcal}(\bpol,q^{\Kcal})]]+\op(n^{-1}h^{-H})\\
    &= \frac{1}{nh^{H}}\left\{\Omega^H_2(k)\E_{\tau}\left[\prod^H_{i=1}\left\{\frac{1}{\bpol(\tau_i(s_i)\mid s_i)} \right\}\var[r_H|s_H,\tau(s_H)] \right]+o(1)\right\}. 
\end{align*}

\subsection{Proof of Theorem \texorpdfstring{\ref{thm:rl_ope2}}{a}}

~
\paragraph{Replacing estimators with true functions}

We define 
$$\phi^{\Kcal}(w^{\Kcal},\bpol, q^{\Kcal})=v^{\Kcal}_0+\sum_{t=1}^H \frac{w^{\Kcal}_tK_h(a_t-\tau_t(s_t)}{\bpol(a_t|s_t)}\{r_t-q^{\Kcal}_t+v^{\Kcal}_{t+1}\}.$$
Here, we prove that nuisance estimators can be replace with true functions:
\begin{align*}\ts
    \P_{\cu_1}[\phi^{\Kcal}(\hat w^{\Kcal*1},\hat \pi^{b\Kcal*1}, \hat q^{\Kcal*1})|\cu_2] +  \P_{\cu_2}[\phi^{\Kcal}(\hat w^{\Kcal*2},\hat \pi^{b\Kcal*2}, \hat q^{\Kcal*2})|\cu_2] &=    \E[\phi^{\Kcal}(w^{\Kcal},\bpol,q^{\Kcal})]+\op((nh)^{-1/2}).
\end{align*}
Then, what we have to prove is 
\begin{align}\ts
   \E[ \phi^{\Kcal}(\hat w^{\Kcal}, \hat \pi^b, \hat q^{\Kcal})-\phi^{\Kcal}(w^{\Kcal},\bpol, q^{\Kcal}) |\cu_2 ]  &=\op((nh)^{-1/2}), \label{eq:first_need}\\
    \E[ \{\phi^{\Kcal}(\hat w^{\Kcal}, \hat \pi^b, \hat q^{\Kcal})-\phi^{\Kcal}(w^{\Kcal},\bpol, q^{\Kcal}) \}^2 |\cu_2] &=\op((h)^{-1}) \label{eq:second_need}. 
\end{align}
By defining $\tilde w^{\Kcal}_t(s_t,a_t)=w^{\Kcal}_t(s_t)/\bpol(a_t|s_t)$, \cref{eq:first_need} is proved by 
{\small 
\begin{align}\ts
    &  \E[ \phi^{\Kcal}(\hat w^{\Kcal}, \hat \pi^b, \hat q^{\Kcal})-\phi^{\Kcal}(w^{\Kcal},\bpol, q^{\Kcal}) |\cu_1 ] \nonumber \\
    &=\E[\sum_{t=1}^H  (\hat {\tilde w}^{\Kcal}_t(S_t,A_t) -\tilde w^{\Kcal}_t(S_t,A_t))K_h((A_t-\tau(S_t))h^{-1})(-\hat q^{\Kcal}_t(S_t,A_t) +q^{\Kcal}_t(S_t,A_t) +\hat v^{\Kcal}_{t+1}(S_{t+1}) -v^{\Kcal}_{t+1}(S_{t+1}))]\label{eq:operl_1}\\
    &=M_2(k)\E[\sum_{t=1}^H \braces{\hat {\tilde w}^{\Kcal}_t(S_t,\tau(S_t)) -\tilde w^{\Kcal}_t(S_t,\tau(S_t))}\braces{-\hat q^{\Kcal}_t(S_t,\tau(S_t)) +q^{\Kcal}_t(S_t,\tau(S_t))}\bpol(\tau(S_t)|S_t)] \label{eq:operl_2} \\
    &+M_2(k)\E[\sum_{t=1}^H \braces{\hat {\tilde w}^{\Kcal}_t(S_t,\tau(S_t)) -\tilde w^{\Kcal}_t(S_t,\tau(S_t))}(\hat v^{\Kcal}_{t+1}(S_{t+1}) -v^{\Kcal}_{t+1}(S_{t+1}))\bpol(\tau(S_t)|S_t)]+\op(h^2)\nonumber  \\
    &=M_2(k)\E[\sum_{t=1}^H \left\{\hat {\tilde w}^{\Kcal}_t(S_t,\tau(S_t)) -\tilde w^{\Kcal}_t(S_t,\tau(S_t))\right\}\left\{-\hat q^{\Kcal}_t(S_t,\tau(S_t)) +q^{\Kcal}_t(S_t,\tau(S_t))\right\}\bpol(\tau(S_t)|S_t)] \label{eq:operl_3}  \\
    &+M_2(k)\E[\sum_{t=1}^H \left\{\hat {\tilde w}^{\Kcal}_t(S_t,\tau(S_t)) -\tilde w^{\Kcal}_t(S_t,\tau(S_t))\right\}\left\{\hat q^{\Kcal}_{t+1}(S_{t+1},\tau(S_{t+1})) -q^{\Kcal}_{t+1}(S_{t+1},\tau(S_{t+1}))\right\}\bpol(\tau(S_t)|S_t)]+\op(h^2) \nonumber \\
    &=M_2(k)\sum_{t=1}^H \|\hat {\tilde w}^{\Kcal}_t-\tilde w^{\Kcal}_t\|_{\infty}\braces{\|\hat q^{\Kcal}_t -q^{\Kcal}_t\|_{\infty}+\|\hat q^{\Kcal}_{t+1} -q^{\Kcal}_{t+1}\|_{\infty}}+\op(h^2)\label{eq:operl_4}  \\
    &=\op(n^{-1/2}h^{-1/2}).  \label{eq:operl_5} 
\end{align}
}
Here, from \eqref{eq:operl_1} to \eqref{eq:operl_2}, 
\begin{align*}
   \max\{\|\hat \pi^{b}- \pi^b\|_{1,\infty},\|\hat w^{\Kcal}_j- w^{\Kcal}_j\|_{\infty},\|\hat q^{\Kcal}_j-q^{\Kcal}_j\|_{1,\infty}\}=\op(1). 
\end{align*}
From \eqref{eq:operl_2} to \eqref{eq:operl_3}, $v^{\Kcal}_t(s)=q^{\Kcal}_t(s,\tau(s))+o(1)$. From  \eqref{eq:operl_3} to  \eqref{eq:operl_4}, we use 
$$\max\{\|\hat \pi^{b}- \pi^b\|_{\infty},\|\hat w^{\Kcal}_j- w^{\Kcal}_j\|_{\infty}\}\|\hat q^{\Kcal}_j-q^{\Kcal}_j\|_{\infty}=\op(n^{-1/2}h^{-1/2})$$. From \eqref{eq:operl_4} to \eqref{eq:operl_5}, we use $nh^5=\bigO(1)$. \cref{eq:second_need} is proved by 
\begin{align*}\ts
       &  \E[ \{\phi^{\Kcal}(\hat w^{\Kcal}, \hat \pi^b, \hat q^{\Kcal})-\phi^{\Kcal}(w^{\Kcal},\bpol, q^{\Kcal}) \}^2 |\cu_1] \\
        &= h^{-1}\max\{\|\hat q^{\Kcal}_t(S_t,\tau(S_t))-q^{\Kcal}_t(S_t,\tau(S_t)) \|^2_2, \|\hat {\tilde w}^{\Kcal}(S_t,\tau(S_t))-{\tilde w^{\Kcal}}(S_t,\tau(S_t))  \|^2_2  \}=\op(h^{-1}).
\end{align*}
~
\paragraph{Calculation of bias and variance term}

Bias and variance terms are bounded as follows. 

\textbf{Bias part}

Bias is the same as the proof of \cref{thm:rl_ope}. It is reduced to 
\begin{align*}\ts
    0.5h^2M_2(k)\sum_{t=1}^{H}\left\{\E_{\tau}[c^{(2)}(s_t,\tau_t(s_t)) ] +\sum_{j=1}^{t-1} \E_{\tau}\left[\frac{r_jp^{(2)}(s_{j+1}\mid s_j,\tau_j(s_j))}{p(s_j\mid s_j,\tau_j(s_j))}\right ] \right\}+\op(n^{-1/2}h^{-1/2}). 
\end{align*}
where $c(s,a)=\E[r|s,a]$.

\textbf{Variance part}

The variance is 
{\small 
\begin{align}\ts
    &\mathrm{var}[\P_n[\phi^{\Kcal}(w^{\Kcal},\bpol,q^{\Kcal})]] \nonumber =\frac{1}{n}\mathrm{var}[\phi^{\Kcal}(w^{\Kcal},\bpol,q^{\Kcal})]] \nonumber \\
     &=\frac{1}{n}\sum_{t=0}^{H} \E\left[\left\{\E\bracks{\prod^t_{i=1}\frac{h^{-1}k(h^{-1}(A_i-S_i))}{\bpol(A_i \mid S_i)}|S_t}\right\}^2\left\{\frac{h^{-1}k(h^{-1}(A_t-\tau(S_t)))}{\bpol(A_t\mid S_t)}\right\}^2\mathrm{\var}[R_t+v^{\Kcal}(S_{t+1})\mid A_t,S_t] \right]. \label{eq:variance3} 
\end{align}
}
First, we have
\begin{align*}\ts
   &\E\left[\prod^{t-1}_{i=1}\frac{h^{-1}k(h^{-1}(A_i-\tau_i(S_i)))}{\bpol(A_i \mid S_i)}\mid S_t \right]\\
    &= \int  \prod^{t-1}_{i=1}\frac{h^{-1}k(h^{-1}(a_i-\tau_i(s_i)))}{\bpol(a_i \mid s_i)}\left\{\prod_{i=1}^{t-1} p(s_{i+1}\mid s_{i},a_{i})\bpol(a_i\mid s_i)\right\}\frac{p_1(s_1)}{p_{\bpol}(s_t)}\rD (h_{a_{t-1}})\\
    &= \int  \prod^{t-1}_{i=1}k(u_i)\prod_{i=1}^{t-1} p(s_{i+1}\mid s_{i},\tau_i(s_i)+hu_i)\frac{p_1(s_1)}{p_{\bpol}(s_t)}\rD (h^{u}_{a_{t-1}})\\
    &= \braces{\int  \prod^{t-1}_{i=1}k(u_i)\prod_{i=1}^{t-1} p(s_{i+1}\mid s_{i},\tau_i(s_i))\frac{p_1(s_1)}{p_{\bpol}(s_t)}\rD (h^{u}_{a_{t-1}})}+o(1)= \frac{p_{\tau_{\theta}}(s_t)}{p_{\bpol}(s_t)}+o(1).
\end{align*}

Therefore, $\forall f$ we have 
\begin{align*}\ts
& \E\left[ \left\{\E\left[\prod^t_{i=1}\frac{h^{-1}k(h^{-1}(A_i-\tau_i(S_i)))}{\bpol(A_i \mid S_i)}\mid S_t \right]\right\}^2 \left\{\frac{h^{-1}k(h^{-1}(A_t-\tau(S_t)))}{\bpol(A_t\mid S_t)}\right\}^2 f(S_t,A_t)\right] \\  
&= \E \left[ \braces{\frac{p_{\tau_{\theta}}(S_t)}{p_{\bpol}(S_t)}+\smallo(1)}^2\left\{\frac{h^{-1}k(h^{-1}(A_t-\tau(S_t)))}{\bpol(A_t\mid S_t)}\right\}^2f(S_t,A_t)\right]+o(1) \\  
&=\frac{1}{h} \braces{\int  \frac{p^2_{\tau_{\theta}}(s_t)}{p_{\bpol}(s_t)}  \left\{\frac{k(u_t)}{\bpol(\tau_t(s_t)+u_t\mid s_t)}\right\}^2f(s_t,a_t)\bpol(\tau_t(s_t)+u_t\mid s_t)\rD (s_t,u_t)+o(1)} \\
&=\frac{1}{h} \braces{\Omega_2(k)\int  \frac{p^2_{\tau_{\theta}}(s_t)}{p_{\bpol}(s_t)\bpol(\tau_t(s_t)\mid s_t)}f(s_t,\tau_t(s_t))\rD (s_t)+o(1)}.  
\end{align*}

In addition, noting
\begin{align*}\ts
  v^{\Kcal}_{t}(s_{t}) &=\int h^{-1}k(h^{-1}(a_t-\tau_t(s_t)))q^{\Kcal}_t(s_t,a_t)\rD (a_t)\\
   &=\int k(u_t)q^{\Kcal}_t(s_t,\tau_t(s_t))\rD (u_t)+o(1)=q^{\Kcal}_t(s_t,\tau_t(s_t))+o(1),
\end{align*}
by induction, we have $v^{\Kcal}_{t}(s_{t})=v^{\epol}_t(s_t)+o(1)=q^{\epol}_t(s_t,\tau_t(s_t))+o(1)$. Then, noting 
\begin{align*}\ts
 \var[R_t+v^{\Kcal}_{t+1}(S_{t+1})\mid S_t=s_t,A_t=a_t]= \var[R_t+v^{\epol}_{t+1}(S_{t+1})\mid S_t=s_t,A_t=a_t]+o(1). 
\end{align*}

Therefore, the variance term is 
\begin{align*}\ts
&\mathrm{var}[\P_n[\phi^{\Kcal}(\hat w^{\Kcal},\hat \pi^b,\hat q^{\Kcal})]]= \mathrm{var}[\P_n[\phi^{\Kcal}(w^{\Kcal},\bpol,q^{\Kcal})]] + \op(n^{-1}h^{-1})\\
&=\frac{1}{nh}\sum_{t=1}^{H} \E_{\tau_{\theta}}\left[\frac{p_{\tau_{\theta}}(s_t)}{p_{\bpol}(s_t)\bpol(\tau_t(s_t)\mid s_t)}   \var[r_t+v^{\epol}_{t+1}(s_{t+1})\mid s_t,\tau_{\theta}(s_t)] \right] \Omega_2(k)+\op(1/nh). 
\end{align*}

\paragraph{Order of main constants in the bias and variance terms}

Bias and variance terms are upper-bounded as follows. 

\textbf{Bound of \texorpdfstring{$B_H$}{b}}

\begin{align*}\ts
    & \sum_{t=1}^{H}\left\{\E_{\tau}\left[\frac{r_t p^{(2)}(r_t|s_t,\tau_t(s_t))}{ p(r_t|s_t,\tau_t(s_t))} \right] +\sum_{j=1}^{t-1} \E_{\tau}\left[\frac{r_{t}p^{(2)}(s_{j+1}\mid s_j,\tau_j(s_j))}{p(s_{j+1}\mid s_j,\tau_j(s_j))}\right ] \right\} \\
    & \leq  R_{\max}\braces{HG_1+\frac{H(H-1)}{2}G_1}. 
\end{align*}
we used an argument:
\begin{align*}\ts
    |\E_{\tau}\left[\frac{r_t p^{(2)}(r_t|s_t,\tau_t(s_t))}{ p(r_t|s_t,\tau_t(s_t))} \right]|&\leq \E_{\tau}\left[\frac{r_t |p^{(2)}(r_t|s_t,\tau_t(s_t))|}{ p(r_t|s_t,\tau_t(s_t))} \right]\leq R_{\max}\E_{\tau}\left[\frac{ |p^{(2)}(r_t|s_t,\tau_t(s_t))|}{ p(r_t|s_t,\tau_t(s_t))} \right]\\ 
    &=  R_{\max}\int |p^{(2)}(r_t|s_t,\tau_t(s_t))|p_{\tau}(s_t)\rD (r_t,s_t)\\
    &\leq  R_{\max}\|\int |p^{(2)}(r_t|s_t,\tau_t(s_t))|\rD r_t\|_{\infty} \leq R_{\max}G^{(2)}_1, \\
 |\E_{\tau}\left[\frac{r_t p^{(2)}(s_{j+1}|s_j,\tau_j(s_j))}{ p(s_{j+1}|s_j,\tau_j(s_j))} \right]|&\leq  \E_{\tau}\left[|\frac{r_t p^{(2)}(s_{j+1}|s_j,\tau_j(s_j))}{ p(s_{j+1}|s_j,\tau_j(s_j))} |\right]\leq R_{\max}\E_{\tau}\left[\frac{|p^{(2)}(s_{j+1}|s_j,\tau_j(s_j))|}{ p(s_{j+1}|s_j,\tau_j(s_j))}\right]\\
     & =R_{\max}\int |p^{(2)}(s_{j+1}|s_j,\tau_j(s_j))|p_{\tau}(s_j)\rD (s_{j+1},s_j)\\
     & \leq R_{\max}\|\int |p^{(2)}(s_{j+1}|s_j,\tau_j(s_j))|p_{\tau}(s_j)\rD s_{j+1}\|_{\infty}\leq R_{\max}G^{(2)}_2. 
\end{align*}

\textbf{Bound of $V_H$}
\begin{align*}\ts
    &\sum_{t=1}^{H} \E_{\tau_{\theta}}\left[\frac{p_{\tau_{\theta}}(s_t)}{p_{\bpol}(s_t)\bpol(\tau_t(s_t)\mid s_t)}\mathrm{var}[r_t+v^{\epol}_{t+1}(s_{t+1})\mid s_t,\tau_{\theta}(s_t) ] \right]  \\ 
    &\leq \sum_{t=1}^{H} \E_{\tau_{\theta}}\left[C_1C_2\mathrm{var}[r_t+v^{\epol}_{t+1}(s_{t+1})\mid s_t,\tau_{\theta}(s_t) ] \right]  \\ 
    &=C_1C_2 \var_{\tau_{\theta}}[\sum_{t=1}^{H} r_t]\leq C_1C_2R^2_{\max}H^2. 
\end{align*}

\subsection{Proof of \texorpdfstring{\cref{thm:gradient_est}}{a}}

~
\paragraph{Replacing estimators with true functions}

Here, we prove that nuisance estimators can be replace with true functions:
\begin{align*}\ts
    &\P_{\cu_1}[\psi^{\Kcal}(\hat w^{\Kcal*1}, \hat q^{\Kcal*1},\hat d^{w^{\Kcal*1}},\hat d^{q^{\Kcal*1}},\hat \pi^b)|\cu_2] +  \P_{\cu_2}[\psi^{\Kcal}(\hat w^{\Kcal*2}, \hat q^{\Kcal*2}, ,\hat d^{w^{\Kcal*2}},\hat d^{q^{\Kcal*2}},\hat \pi^b)|\cu_1] \\
    &=   \E[\phi^{\Kcal}(w^{\Kcal},q^{\Kcal},d^{w^{\Kcal}}, d^{q^{\Kcal}},\pi^b)]+\op((nh^3)^{-1/2}).
\end{align*}
Then, what we have to prove is 
\begin{align}\ts
   \E[ \psi^{\Kcal}(\hat w^{\Kcal}, \hat q^{\Kcal},\hat d^{w^{\Kcal}},\hat d^{q^{\Kcal}},\hat \pi^b)-\psi^{\Kcal}(w^{\Kcal}, q^{\Kcal},d^{w^{\Kcal}}, d^{q^{\Kcal}},\pi^b) |\cu_2]  &=\op((nh^3)^{-1/2}) ,\label{eq:first_need_}\\
    \E[ \{\psi^{\Kcal}(\hat w^{\Kcal}, \hat q^{\Kcal},\hat d^{w^{\Kcal}},\hat d^{q^{\Kcal}},\hat \pi^b)-\psi^{\Kcal}(w^{\Kcal}, q^{\Kcal},d^{w^{\Kcal}}, d^{q^{\Kcal}},\pi^b)\}^2 |\cu_2] &=\op((h)^{-1}) \label{eq:second_need_}. 
\end{align}

\cref{eq:first_need_} is proved by 
{\small 
\begin{align*}\ts
    &  \E[ \psi^{\Kcal}(\hat w^{\Kcal}, \hat q^{\Kcal},\hat d^{w^{\Kcal}},\hat d^{q^{\Kcal}},\hat \pi^b)-\psi^{\Kcal}(w^{\Kcal}, q^{\Kcal},d^{w^{\Kcal}}, d^{q^{\Kcal}},\pi^b) |\cu_1 ]  \\
    &=\E[\sum_{t=1}^H  \{\hat d^{{w}^{\Kcal}}_t(S_t)\hat \eta(S_t,A_t) -d^{w^{\Kcal}}_t(S_t)\eta(S_t,A_t))\}K_h((A_t-\tau(S_t)))(-\hat q^{\Kcal}_t(S_t,A_t) +q^{\Kcal}_t(S_t,A_t) +\hat v^{\Kcal}_{t+1}(S_{t+1}) -v^{\Kcal}_{t+1}(S_{t+1}))] \\
    &+\E[\sum_{t=1}^H  \braces{\hat {w}^{\Kcal}_t(S_t)\hat \eta(S_t,A_t) -w^{\Kcal}_t(S_t)\eta(S_t,A_t)}K_h((A_t-\tau(S_t)))(-d^{\hat q^{\Kcal}}_t(S_t,A_t) +d^{q^{\Kcal}}_t(S_t,A_t) +d^{\hat v^{\Kcal}}_{t+1}(S_{t+1}) -d^{v^{\Kcal}}_{t+1}(S_{t+1}))] \\
    &-\E[\sum_{t=1}^H   \braces{\hat {w}^{\Kcal}_t(S_t)\hat \eta(S_t,A_t) -w^{\Kcal}_t(S_t)\eta(S_t,A_t)}\times \\ 
    &h^{-2} k^{(1)}((A_t-\tau(S_t))h^{-1})\nabla \tau(S_t)(-\hat q^{\Kcal}_t(S_t,A_t) +q^{\Kcal}_t(S_t,A_t) +\hat v^{\Kcal}_{t+1}(S_{t+1}) -v^{\Kcal}_{t+1}(S_{t+1}))] \\
    &=\op(h^{-2})+\sum_{t=1}^H  \max\{\|\hat w^{\epol}_t-w^{\epol}_t \|_{\infty},\|\hat \eta_t-\eta_t  \|_{1,\infty},\|\hat d^{{w}^{\Kcal}}_t -d^{w^{\Kcal}}_t\|_{\infty}\}\\
    &\,\,\,\times \max\{\|\hat q_t-q_t \|_{1,\infty},\|\hat q_{t+1}-q_{t+1}  \|_{1,\infty},\|\hat d^{{q}^{\Kcal}}_t -d^{q^{\Kcal}}_t\|_{\infty},\|\hat d^{{q}^{\Kcal}}_{t+1} -d^{q^{\Kcal}}_{t+1}\|_{\infty}\}\\
    &=\op(n^{-1/2}h^{-3/2}). 
\end{align*}
}
\cref{eq:second_need_} is similarly proved. 

~
\paragraph{Bias part}

First, we have 
\begin{align*}\ts
    \E[\P_n[\psi(w^{\Kcal}, q^{\Kcal},d^{w^{\Kcal}}, d^{q^{\Kcal}},\pi^b)]]=\nabla\E[\sum_t \lambda^{\Kcal}_t r_t]-\nabla J(\theta)=\nabla\{\sum_t \E[\{\lambda^{\Kcal}_t-\lambda_t\} r_t]\}.
\end{align*}   
Here, we use a doubly robust property of $\phi$. Then, the above is equal to 
{\small 
\begin{align*}\ts   
    &-\sum_t\left\{\int c(s_t,a_t)\sum_{i=1}^{t}\frac{k^{(1)}((a_i-\tau({s_i}))h^{-1})}{h^2}\nabla \tau_i(s_i) \prod_{j\neq i}^t \frac{k((a_j-\tau({s_j}))h^{-1}) }{h}\prod_{l=1}^{t}p(s_l \mid s_{l-1},a_{l-1}) \}\rD (h_{a_t})-\E[\lambda_t r_t]\right\}\\
     &=-\sum_{t=1}^H \left\{\int c(s_t,\tau_t(s_t)+hu_t)\sum_{i=1}^{t}h^{-1}k^{(1)}(u_i)\nabla \tau_i(s_i) \prod_{j\neq i}^t k(u_j) \prod_{l=1}^{t}p(s_l \mid s_{l-1},\tau(s_{l-1})+hu_l) \rD (h^{u}_{a_t})-\E[\lambda_t r_t]\right\} \\
      &=\sum_{t=1}^H \{\int c^{(1)}_t(s_t,\tau_t(s_t)+hu_t)\nabla \tau_t(s_t)\prod_{j=1}^{t}k(u_j) \prod_{i=1}^{t}p(s_l \mid s_{i-1},\tau(s_{i-1})+hu_i) \rD (h^{u}_{a_t})+ \\
      &+\int c(s_t,\tau_t(s_t)+hu_t) \sum_{i=1}^t \nabla \tau_i(s_i)\frac{p^{(1)}(s_i \mid s_{i-1},\tau(s_{i-1})+hu_i)}{p(s_i \mid s_{i-1},\tau(s_{i-1})+hu_i) } \prod_{j=1}^{t}k(u_j)\prod_{l=1}^{t}p(s_l \mid s_{l-1},\tau(s_{l-1})+hu_l) \rD (h^{u}_{a_t}) \\
      &-\E[\lambda_t r_t]\}. 
\end{align*}
}
This is equal to 
{\small 
\begin{align*}
     &0.5h^2M_2(\Kcal) \sum_t \{\int c^{(3)}_t(s_t,\tau_t(s_t)) \prod_{i=1}^{t}p(s_l \mid s_{i-1},\tau(s_{i-1})) \rD (h_{s_t})+ \\
    &\int c^{(1)}_t(s_t,\tau_t(s_t))\nabla \tau_t(s_t) \sum_{j=2}^{t}\frac{p^{(2)}(s_j \mid s_{j-1},\tau(s_{j-1})) }{p(s_l \mid s_{i-1},\tau(s_{i-1}))}\prod_{l=1}^{t}p(s_l \mid s_{i-1},\tau(s_{i-1})) \rD (h^{-u}_{s_t})+ \\
     &+\int c^{(2)}_t(s_t,\tau_t(s_t)) \sum_{i=2}^t \nabla \tau_i(s_i)\frac{p^{(1)}(s_i \mid s_{i-1},\tau(s_{i-1}))}{p(s_i \mid s_{i-1},\tau(s_{i-1})) }\prod_{l=1}^{t}p(s_l \mid s_{l-1},\tau(s_{l-1})) \rD (h^{-u}_{s_t}) \\
     &+\int c(s_t,\tau_t(s_t)) \sum_{i=2}^t \nabla \tau_i(s_i)\left\{\frac{p^{(3)}(s_i \mid s_{i-1},\tau(s_{i-1}))}{p(s_i \mid s_{i-1},\tau(s_{i-1})) }+ \frac{p^{(1)}(s_i \mid s_{i-1},\tau(s_{i-1}))}{p(s_i \mid s_{i-1},\tau(s_{i-1})) }\sum_{j\neq i}\frac{p^{(2)}(s_j \mid s_{j-1},\tau(s_{j-1}))}{p(s_j \mid s_{j-1},\tau(s_{j-1})) }\right\} \\
     &\,\,\,\,\times \prod_{l=1}^{t}p(s_l \mid s_{l-1},\tau(s_{l-1})) \rD (h^{-u}_{s_t})\\
&= 0.5h^2M_2(k)\tilde B_H. 
\end{align*}
}
Here, $h^{-u}_{a_t}=\{s_1,s_2,\cdots \}$. 
In the end, $\tilde B_H$ is equal to 
{\small 
\begin{align*}\ts 
 &\sum_{t=1}^H\braces{\E_{\tau}\left[r_t\frac{p^{(3)}(r_t|s_t,\tau_t(s_t))}{p(r_t|s_t,\tau_t(s_t))} \nabla \tau_t(s_t)\right]+\sum_{j=1}^{t-1}\E_{\tau}\left[r_t\frac{p^{(2)}(r_t|s_t,\tau_t(s_t))p^{(1)}(s_{j+1}|s_{j},\tau_j(s_j))}{p(r_t|s_t,\tau_t(s_t))p(s_{j+1}|s_j,\tau_j(s_j))} \nabla \tau_j(s_j)\right] }+\\
 &+\sum_{t=1}^H \sum_{j=1}^{t-1}\E_{\tau}\left[r_t\frac{p^{(1)}(r_t|s_t,\tau_t(s_t))p^{(2)}(s_{j+1}|s_j,\tau_j(s_j))}{p(r_t|s_t,\tau_t(s_t))p(s_{j+1}|s_j,\tau_j(s_j))} \nabla \tau_t(s_t)\right]+ \\
 &+\sum_{t=1}^H \braces{\sum_{j=1}^{t-1}\E_{\tau}\left[r_t\frac{p^{(3)}(s_{j+1}|s_j,\tau_j(s_j))}{p(s_{j+1}|s_j,\tau_j(s_j))} \nabla \tau_j(s_j)\right]+\sum_{j\neq i}^{t-1}\E_{\tau}\left[r_t\frac{p^{(1)}(s_{i+1}|s_i,\tau_i(s_i))p^{(2)}(s_
 {j+1}|s_j,\tau_j(s_j))}{p(s_{i+1}|s_i,\tau_i(s_i))p(s_{j+1}|s_j,\tau_j(s_j))} \nabla \tau_i(s_i)\right]}.  
\end{align*}
}
The operator norm of $\otimes \tilde B_H$ is upper bounded by
\begin{align*}\ts
    R^2_{\max}\Upsilon^2\braces{HG^{(3)}_1+\frac{H(H-1)}{2}\{G^{(2)}_1G^{(1)}_2+G^{(1)}_1G^{(2)}_2+G^{(3)}_2 \}+\frac{H(H-1)(H-2)}{3} G^{(1)}_2G^{(2)}_2 }^2. 
\end{align*}
For example, 
\begin{align*}
    \left \|\otimes \E_{\tau}\left[r_t\frac{p^{(3)}(s_{j+1}|s_j,\tau_j(s_j))}{p(s_{j+1}|s_j,\tau_j(s_j))} \nabla \tau_j(s_j)\right]\right\|_{\oper}&\leq     \left \|\otimes \E_{\tau}\left[\frac{|r_tp^{(3)}(s_{j+1}|s_j,\tau_j(s_j))|}{p(s_{j+1}|s_j,\tau_j(s_j))} \nabla \tau_j(s_j)\right]\right\|_{\oper}\\
    &\leq R^2_{\max}G^{2(3)}_2\left \|\otimes \E_{\tau}\left[\nabla \tau_j(s_j) \right]\right \|\|_{\oper}\\
        &\leq R^2_{\max}G^{2(3)}_2\left \|\E_{\tau} \left[\otimes \nabla \tau_j(s_j) \right]\right \|_{\oper}\leq R^2_{\max}G^{2(3)}_2\Upsilon^2. 
\end{align*}

\paragraph{Variance part}

The variance part is calculated as 
\begin{align*}\ts
&\frac{1}{n}\sum_{h=1}^H\E\left[\frac{w^{2\Kcal}_t(S_t)k^{2(1)}(\{(A_t-\tau(S_t)\}h^{-1})}{h^4\pi^{2b}_t(A_t|S_t)}\var[R_t+v^{\Kcal}_{t+1}(S_{t+1}) |S_t,A_t]\otimes\nabla \tau(S_t)\right]+\op(h^3/n)\\
&=\frac{1}{nh^3}\sum_{h=1}^H\int k^{2(1)}(u_h)\rD (u_h) \E\left[\frac{w^{2\Kcal}(S_t) }{\pi^{2b}_t(\tau(S_t)|S_t)}\var[R_t+v^{\epol}_{t+1}(S_{t+1}) |S_t,\tau(S_t)]\otimes\nabla \tau(S_t)\right]+\op(h^3/n) \\
&=\frac{\Omega^{(1)}_2(k)}{nh^3}\sum_{h=1}^H \E_{\tau}\left[\frac{w^{\epol}(s_t) }{\pi^{b}_t(\tau_t(s_t)|s_t)}\var[r_t+v^{\epol}_{t+1}(s_{t+1}) |s_t,\tau_t(s_t)]\otimes\nabla \tau_t(s_t)\right]+\op(h^3/n) \\
&=\frac{\Omega^{(1)}_2(k)}{nh^3}\tilde V_H+\op(h^3/n) .
\end{align*}
Then, the operator norm of $\tilde V_H$ is upper bounded as
\begin{align*}\ts 
&\left \|\sum_{t=1}^H \E_{\tau}\left[\frac{w^{\epol}(s_t) }{\pi^{b}_t(\tau_t(s_t)|s_t)}\var[r_t+v^{\epol}_{t+1}(s_{t+1}) |s_t,\tau_t(s_t)]\otimes\nabla \tau_t(s_t)\right]\right\|_{\oper}\\
&\leq \|C_1C_2\sum_{t=1}^H \E_{\tau}\left[\var[r_t+v^{\epol}_{t+1}(s_{t+1}) |s_t,\tau_t(s_t)]\otimes\nabla \tau_t(s_t)\right]\|_{\oper}\\
&\leq \|C_1C_2\|\otimes \nabla \tau(s)\|_{\infty} \sum_{t=1}^H \E_{\tau}\left[\var[r_t+v^{\epol}_{t+1}(s_{t+1}) |s_t,\tau_t(s_t)]\right]\|_{\oper}\\
&\leq \|C_1C_2\|\otimes \nabla \tau(s)\|_{\infty} R^2_{\max}H^2]\|_{\oper} \leq C_1C_2R_{\max}H^2\Upsilon^2. 
\end{align*}

\section{Different Representation of Theorem \ref{thm:ope_final} }\label{sec:probability}

\begin{theorem}
\label{thm:ope_final2}
Suppose for $i=1,2$,
$\|\hat \pi^{b,[i]}(A \mid S)- \pi^b(A\mid S)\|_{1,\infty}=\op(1)$, $\|\hat q^{[i]}(S,A)- q(S,A)\|_{1,\infty}=\op(1)$, $\|\hat \pi^{b,[i]}(A \mid S)- \pi^b(A\mid S)\|_{\infty}\|\hat q^{[i]}(S,A)- q(S,A)\|_{\infty}=\op((nh)^{-1/2})$, $nh^5=\bigO(1)$, $nh\to \infty$, that $\bpol(a\mid s),\,q(s,a)$ are twice continuously differentiable wrt $a$ for almost all $s$, and that $\hat \pi^{b,[i]},\hat q^{[i]}$ are uniformly bounded by a constant.
Then, for any small $\epsilon,
\delta>0$, there exists $N_{\epsilon,\delta}$, and for all $n\geq N_{\epsilon,\delta}$, on some event $A_n$ s.t. $P(A_n)>1-\epsilon$, the bias and variance of $\hat J^{\Dcal}$ are $|\E[\hat J^{\Dcal}|A_n]-J-0.5M_2(k)h^2B|<\delta /(nh)^{-1/2},\,|\var[\hat J^{\Dcal}|A_n]-\frac{\Omega_2(k)}{nh}|<\delta^2 /(nh)$, where
\begin{align*}\ts
  B=\E[q^{(2)}(S,\tau(S))+2q^{(1)}(S,\tau(S))\pi^{b(1)}(\tau(S)|S)/\pi^{b}(\tau(S)|S)],\,V=\E\left[\frac{\mathrm{var}[R\mid S,\tau(S)]}{\pi^b(\tau(S)\mid S) } \right]. 
\end{align*}
If additionally $\hat\pi^{b,[i]}(a \mid s),\,\hat q^{[i]}(s,a)$ are twice continuously differentiable wrt $a$, then the same holds for $\hat J^{\Kcal}$ with $B=\E[q^{\branormal{2}}(S,\tau(S))]$ and the same $V$ as the above. In both cases, setting $h=\Theta(n^{-1/5})$ yields the minimal MSE of order $\bigO(n^{-4/5})$.
\end{theorem}

We prove \cref{thm:ope_final2}. To do that, we prove \cref{thm:ope_middle2}. The rest of the proof is the same the that of \cref{thm:ope_final}.

\begin{theorem}\label{thm:ope_middle2}
In the case $\Kcal$, assuming (b), we have 
\begin{align}\ts
\label{eq:replace2}
  \hat J= \E_{n}\left[ \frac{K_h(A-\tau(S))\{R-f_{1}\}}{f_2}+f_3\right]+\op((nh)^{-1/2}).
\end{align}
In the case $\Dcal$, the above \eqref{eq:replace2} also holds assuming (a).
\end{theorem} 

\subsubsection{Proof of \texorpdfstring{\cref{thm:ope_middle2}}{a}, Case \texorpdfstring{$\Kcal$}{a}}

Let us define 
\begin{align*}\ts
    \phi_1(s,a,r;q,\bpol)= \frac{K_h(a-\tau(s))\{r-q(s,a) \}}{\bpol(a\mid s)}+v(s),v(s)=\int K_h(a-\tau(s))q(s,a)\rD (a). 
\end{align*}
As in a standard argument \citep{ChernozhukovVictor2018LRSE,KallusUehara2019}, what we have to prove is 
\begin{align}\ts
    \E[\phi_1(S,A,R;\hat q^{*1},\hat \pi^{b*1})-\phi_1(S,A,R;q,\pi^{b})\mid \cu_2]&= \op((nh)^{-1/2}), \label{eq:firstfirst2}\\
    \E[\{\phi_1(S,A,R;\hat q^{*1},\hat \pi^{b*1})-\phi_1(S,A,R;q,\pi^{b})\}^2\mid \cu_2]&= \op(h^{-1}). \label{eq:secondsecond2}
\end{align}
Then, the same argument holds for $\phi_1(S,A,R;\hat q^{*2},\hat \pi^{b*2})-\phi_1(S,A,R;q,\pi^{b})$. The desired statement is concluded since 
\begin{align}\ts
&\P_n[ \phi_1(S,A,R;\hat q,\hat \pi^{b}) ]- \P_n[ \phi_1(S,A,R;q,\pi^{b}) ] \nonumber \\
&= \bG_n[\phi_1(S,A,R;\hat q,\hat \pi^{b})-\phi_1(S,A,R;q,\pi^{b}) \mid \cu_2 ] \label{eq:decomoposion2}\\
&+  \E[\phi_1(S,A,R;\hat q,\hat \pi^{b})-\phi_1(S,A,R;q,\pi^{b})\mid \cu_2]\nonumber \\
&= \op((nh)^{-1/2})+ \op((nh)^{-1/2}). \nonumber 
\end{align}
Especially, \eqref{eq:decomoposion2} is $\op((nh)^{-1/2})$ since for $\epsilon>0$, 
\begin{align*}\ts
   &\P(n^{1/2}h^{1/2}\times \bG_{\cu_2}[\phi_1(S,A,R;\hat q^{*1},\hat \pi^{b^{*1}})-\phi_1(S,A,R;q,\pi^{b})]>\epsilon|\cu_2)\\
   &\leq \E[h\{\phi_1(S,A,R;\hat q^{*1},\hat \pi^{b*1})-\phi_1(S,A,R;q,\pi^{b})\}^2   /\epsilon^2 |\cu_2]=\op(1). 
\end{align*}
Therefore, noting $\P(n^{1/2}h^{1/2}\bG_{\cu_2}[\phi_1(S,A,R;\hat q^{*1},\hat \pi^{b^{*1}})-\phi_1(S,A,R;q,\pi^{b})]>\epsilon|\cu_2)$ is uniformly integrable, 
\begin{align*}\ts
       \P(n^{1/2}h^{1/2}\bG_{\cu_2}[\phi_1(S,A,R;\hat q^{*1},\hat \pi^{b^{*1}})-\phi_1(S,A,R;q,\pi^{b})]>\epsilon|\cu_2)=\op(1)
\end{align*}
implies
\begin{align*}\ts
    \bG_{\cu_2}[\phi_1(S,A,R;\hat q^{*1},\hat \pi^{b^{*1}})-\phi_1(S,A,R;q,\pi^{b})]= \op(n^{-1/2}h^{-1/2}). 
\end{align*}

\paragraph{Proof of \texorpdfstring{\cref{eq:firstfirst2}}{a} and \texorpdfstring{\cref{eq:secondsecond2}}{a}}

In this subsection, we remove $\{*1\}$ for the ease of the notation. To prove \eqref{eq:firstfirst}, we show $$\E\left[K_h(A-\tau(S))\left\{\frac{1}{\hat \pi^b(A|S)}-\frac{1}{\pi^b(A|S)}\right\}\{q(S,A)-\hat q(S,A)\} \mid \cu_2 \right]=\op((nh)^{-1/2}).$$
This is proved by 
{\small 
\begin{align*}\ts
    & \E\left[K_h(A-\tau(S))\left\{\frac{1}{\hat \pi^b(A|S)}-\frac{1}{\pi^b(A|S)}\right\}\{q(S,A)-\hat q(S,A)\} \mid \cu_2 \right]\\
    &=\int \frac{1}{h}k((a-\tau(s))h^{-1}) \left\{\frac{1}{\hat \pi^b(a|s)}-\frac{1}{\pi^b(a|s)}\right\}\{q(s,a)-\hat q(s,a)\}\bpol(a\mid s)p(s)\rD (s,a)\\
     &=\int k(u) \left\{\frac{1}{\hat \pi^b(\tau(s)+uh|s)}-\frac{1}{\pi^b(\tau(s)+uh|s)}\right\}\{q(s,\tau(s)+uh)-\hat q(s,\tau(s)+uh)\}\bpol(\tau(s)+uh\mid s)p(s)\rD (u,s)\\
    &=\int k(u) \left\{\frac{1}{\hat \pi^b(\tau(s)|s)}-\frac{1}{\pi^b(\tau(s)|s)}+uh\braces{-\frac{\hat \pi^{b(1)}(\tau(s)|s)}{\hat \pi^{2b}(\tau(s)|s)}+\frac{\pi^{b(1)}(\tau(s)|s)}{\pi^{2b}(\tau(s)|s)}}+\bigO((uh^2)) \right\}\times\\
    &\{q(s,\tau(s))-\hat q(s,\tau(s))+uh\{q^{(1)}(s,\tau(s))-\hat q^{(1)}(s,\tau(s))\}+\bigO((uh^2))\}\times \\
    &\{\bpol(\tau(s)\mid s)+uh\pi^{b(1)}(\tau(s)\mid s)+\bigO((uh^2))\}p(s)\rD (u,s)\\
     &=M_2(k)\int \left\{\frac{1}{\hat \pi^b(\tau(s)\mid s)}-\frac{1}{\pi^b(\tau(s)\mid s)}\right\}\{q(s,\tau(s))-\hat q(s,\tau(s))\}\bpol(\tau(s)\mid s)p(s)\rD (s)+\bigO(h^2)\times o_{p}(1)\\
     &=\op((nh)^{-1/2}). 
\end{align*}
}
\normalsize
In the last line, we use the assumptions that $q(a,x),\bpol(a|x),\hat q(a,x),\hat \pi^b(a|x)$ are $C^2$-functions wrt actions, and 
\begin{align*}\ts
    &\left \|\frac{1}{\hat \pi^b(A\mid S)}- \frac{1}{\pi^b(A\mid S)} \right\|_{\infty}\|q(S,A)-\hat q(S,A) \|_{\infty}=\op(n^{-1/2}h^{-1/2}),\\
    &\left \|\frac{1}{\hat \pi^b(A)\mid S)}- \frac{1}{\pi^b(A\mid S)} \right\|_{1,\infty}=\op(1),\|\hat q(S,A)-\hat q(S,A)\|_{1,\infty}=\op(1),\\
    &\bigO(h^2)\times o_{p}(1)=\op((nh)^{-1/2}),nh^5=\bigO(1). 
\end{align*}

In addition, \cref{eq:secondsecond2} is proved since 
\begin{align*}\ts
      &\E[\{\phi_1(S,A,R;\hat q^{*1},\hat \pi^{b*1})-\phi_1(S,A,R;q^{},\pi^{b})\}^2\mid \cu_2]\\
    &\lessapprox  \E\left[K_h(A-\tau(S))^2\left\{\frac{1}{\hat \pi^b(A\mid S)}-\frac{1}{\pi^b(A\mid S)}\right\}^2\{q(S,A)-\hat q(S,A)\}^2 \mid \cu_2\right]  \\
    &+\E\left[K_h(A-\tau(S))^2\left\{\frac{1}{\hat \pi^b(A \mid S)}-\frac{1}{\pi^b(A \mid S)}\right\}^2 \{R-q(S,A)\}^2\mid \cu_2 \right] \\
    &+\E\left[\frac{K_h(A-\tau(S))^2}{\pi^b(A\mid S)^2}\{q(S,A)- \hat q(S,A)\}^2\mid \cu_2  \right]+\E\left[\braces{\hat v(S)-v(S)}^2\mid \cu_2 \right]   \\
     &\lessapprox h^{-1}\max\left\{\|\hat q(S,\tau(S))-q(S,\tau(S))\|^2_2, \left \|\frac{1}{\hat \pi^b(\tau(S)|S)}-\frac{1}{\pi^b(\tau(S)|S)}\right\|^2_2  \right\}+\bigO(1)=\op(h^{-1}).
\end{align*}

\subsubsection{Proof of \texorpdfstring{\cref{thm:ope_middle2}}{a}, Case \texorpdfstring{$\Dcal$}{a}}

Essentially, the same proof is seen in \citet{Kyle2019}. For completeness, we also write the proof here with our notation. Let us define 
\begin{align*}\ts
    \phi_2(s,a,r;q,\bpol)= \frac{K_h(a-\tau(s))\{r-q(s,\tau(s)) \}}{\bpol(a\mid s)}+q(S,\tau(S)). 
\end{align*}
As in a standard argument similar to the case $\Kcal$, what we have to prove is 
\begin{align}\ts
    \E[\phi_2(S,A,R;\hat q^{*1},\hat \pi^{b*1})-\phi_2(S,A,R;q^{},\pi^{b})|\cu_2] &=\op((nh)^{-1/2}), \label{eq:firstfirst22}\\
    \E[\{\phi_2(S,A,R;\hat q^{*1},\hat \pi^{b*1})-\phi_2(S,A,R;q^{},\pi^{b})\}^2|\cu_2]&= \op(h^{-1}). \label{eq:secondsecond22}
\end{align}
In this subsection, we remove $\{*1\}$ for the ease of the notation. 

\cref{eq:firstfirst2} is proved since 
\begin{align}\ts
    &\E[\phi_2(S,A,R;\hat q,\hat \pi^b)-\phi_2(S,A,R;q,\pi^b)\mid \cu_2] \nonumber \\ 
    &=\E\left[K_h(A-\tau(S))\left\{\frac{1}{\hat \pi^b(\tau(S)\mid S)}-\frac{1}{\pi^b(\tau(S)\mid S)}\right\}\{q(S,\tau(S))-\hat q(S,\tau(S))\} \mid \cu_2\right]+  \label{eq:first2} \\
    &+\E\left[K_h(A-\tau(S))\left\{\frac{1}{\hat \pi^b(\tau(S)\mid S)}-\frac{1}{\pi^b(\tau(S)\mid S)}\right\} \{R-q(S,\tau(S))\}\mid \cu_2 \right] \label{eq:second2}\\
    &+\E\left[\frac{K_h(A-\tau(S))}{\pi^b(\tau(S)\mid S)}\{q(S,\tau(S))- \hat q(S,\tau(S))\}+\hat q(S,\tau(S))-q(S,\tau(S))\mid \cu_2 \right]    \label{eq:third2} \\
     &=\op((nh)^{-1/2})+\op(1)\times \bigO(h^2)+ \op(1)\times \bigO(h^2)=\op(nh)^{-1/2}.\nonumber 
\end{align}
Here, we use the facts that \eqref{eq:first2} is $\op((nh)^{-1/2})$, \eqref{eq:second2} is $\op(1)\times \bigO(h^2)$, \eqref{eq:third2} is $\op(1)\times \bigO(h^2)$, which we will prove soon. In the last line, we use $nn^5=\bigO(1)$. 
From now on, we prove \eqref{eq:second2} is $\op(1)\times \bigO(h^2)$:
{\small 
\begin{align*}\ts
    & \E\left[K_h(A-\tau(S))\left\{\frac{1}{\hat \pi^b(\tau(S)\mid S)}-\frac{1}{\pi^b(\tau(S)\mid S)}\right\} \{R-q(S,\tau(S))\}\mid \cu_2 \right] \\
    &=\E\left[\left\{\frac{1}{\hat \pi^b(\tau(S)\mid S)}-\frac{1}{\pi^b(\tau(S)\mid S)}\right\}\{\E[K_h(A-\tau(S))q(S,A)\mid S]-K_h(A-\tau(S))q(S,\tau(S))\}\mid \cu_2  \right] \\ 
 &=\E\left[\left\{\frac{1}{\hat \pi^b(\tau(S)\mid S)}-\frac{1}{\pi^b(\tau(S)\mid S)}\right\}\left\{\bigO(h^2)\right\} \mid \cu_2  \right] \\
 &= \op(1)\times \bigO(h^2).
\end{align*}
}
More  specifically,
\begin{align*}
    &\E[K_h(A-\tau(S))\{q(S,A)-q(S,\tau(S))\}\mid S]=\int \frac{1}{h}k\braces{\frac{a-\tau(s)}{h}}\bpol(a|s)\{q(s,a)-q(s,\tau(s))\}\mathrm{d}a\\
    &=\int k(u)\bpol(\tau(s)+uh|s)\{q(s,\tau(s)+uh)-q(s,\tau(s))\}\mathrm{d}u\\
        &=\int k(u)\{\bpol(\tau(s)|s)+\bigO(uh)\}\{uh q^{(1)}(s,\tau(s))+\bigO(h^2) \}\mathrm{d}u=\bigO(h^2). 
\end{align*}
noting $\int uk(u)\mathrm{d}u=0$. Next, we prove \eqref{eq:third} is $\op(1)\times \bigO(h^2)$: 
\begin{align*}\ts
&\E\left[\frac{K_h(A-\tau(S))}{\pi^b(\tau(S)\mid S)}\{q(S,\tau(S))- \hat q(S,\tau(S))\}+\hat q(S,\tau(S))-q(S,\tau(S))\mid \cu_2 \right] \\
&=\E\left[\left\{\frac{K_h(A-\tau(S))}{\pi^b(\tau(S)\mid S)}-1\right\}\{q(S,\tau(S))- \hat q(S,\tau(S))\}\mid \cu_2 \right]\\ 
&=\E\left[\{\bigO(h^2)\}\{q(S,\tau(S))- \hat q(S,\tau(S))\}\mid \cu_2 \right]=\op(1)\times \bigO(h^2). 
\end{align*}

\cref{eq:secondsecond2} is similarly proved as in the case $\Kcal$.

\end{document}

%% file: def.tex
\usepackage[utf8]{inputenc}
\usepackage[T1]{fontenc}
\usepackage{graphicx}
\usepackage{subcaption}
\usepackage{mathtools}
\usepackage{booktabs}
\usepackage{bm}
\usepackage{tcolorbox}
\usepackage{multicol,multirow}
\usepackage[sort]{natbib}
\usepackage{comment}
\newcommand{\cu}{\mathcal{U}}

\newcommand{\traj}{\mathcal{T}}
\usepackage{booktabs}
\newcommand{\ie}{\textit{i.e.}}

\usepackage{enumerate}   
\usepackage{amsmath}
\usepackage{multirow}
\usepackage{bbm}
\usepackage{tikz}

\definecolor{mygreen}{rgb}{0,0.6,0}
\usetikzlibrary{positioning,arrows,decorations,decorations.markings,shadows,positioning,arrows.meta,matrix,fit}

\usetikzlibrary{positioning,arrows}
\usepackage{wrapfig}

\usepackage{amsmath}
\usepackage{appendix}
\usepackage{amssymb}

\usepackage{amsthm}
\usepackage{hyperref}
\usepackage[capitalise]{cleveref}
\Crefname{assumption}{Assumption}{Assumptions}

\theoremstyle{plain}
\newtheorem{theorem}{Theorem}

\newtheorem{corollary}{Corollary}
\theoremstyle{definition}

\newtheorem{remark}{Remark}

\newcommand{\smallo}{{\scriptscriptstyle\mathcal{O}}} %

\newcommand{\bG}{\mathbb{G}}

\newcommand{\CK}{\mathrm{cdr}}
\newcommand{\CD}{\mathrm{cdr}}
\newcommand{\MK}{\mathrm{mpg}}
\newcommand{\MD}{\mathrm{mpg}}

\newcommand{\ch}{\mathcal{H}}
\newcommand{\bigO}{\mathcal{O}} %

\newcommand{\oper}{\mathrm{op}}
\newcommand{\ts}{\textstyle}

\renewcommand{\P}{\mathbb{P}}

\newcommand{\op}{\mathrm{o}_{p}}
\newcommand{\E}{\mathbb{E}}

\newcommand{\pa}{\mathrm{\pa}}

\newcommand{\var}{\mathrm{var}}

\newcommand{\rD}{\mathrm{d}}
\newcommand{\rd}{\mathrm{d}}

\newcommand{\prns}[1]{\left(#1\right)}
\newcommand{\braces}[1]{\left\{#1\right\}}
\newcommand{\bracks}[1]{\left[#1\right]}
\newcommand{\braangle}[1]{\langle#1\rangle}
\newcommand{\branormal}[1]{(#1)}

\newcommand{\Rl}{\mathbb{R}}

\newcommand{\epol}{\pi^{\mathrm{e}}}
\newcommand{\bpol}{\pi^{\mathrm{b}}}

\newcommand{\Kcal}{\mathcal{K}}
\newcommand{\Scal}{\mathcal{S}}
\newcommand{\Acal}{\mathcal{A}}

\newcommand{\Dcal}{\mathcal{D}}

\renewcommand{\eqref}[1]{(\ref{#1})}

\newcommand{\RN}[1]{%
  \textup{\uppercase\expandafter{\romannumeral#1}}%
}

\usepackage{color,graphicx,lscape,longtable}

\def\boxit#1{\vbox{\hrule\hbox{\vrule\kern6pt\vbox{\kern6pt#1\kern6pt}\kern6pt\vrule}\hrule}}

\usepackage{xcolor}
\newcount\Comments  
\newcommand{\kibitz}[2]{\ifnum\Comments=1\textcolor{#1}{#2}\fi}